\documentclass[journal]{IEEEtran}
\usepackage{amsmath,amsfonts}

\usepackage[linesnumbered,ruled]{algorithm2e}
\usepackage{array}
\usepackage{booktabs}
\usepackage[colorlinks=true, linkcolor=blue, citecolor=blue, urlcolor=blue, pdfborder={0 0 0}]{hyperref}
\usepackage[caption=false,font=normalsize,labelfont=sf,textfont=sf]{subfig}
\usepackage{textcomp}
\usepackage{stfloats}
\usepackage{url}
\usepackage{verbatim}
\usepackage{graphicx}
\usepackage{makecell}
\usepackage{bbding}
\usepackage{epstopdf}
\usepackage{changepage}
\usepackage[switch]{lineno}
\usepackage{multirow}
\usepackage{amsthm}
\newtheorem{remark}{Remark}

\newtheorem{theorem}{Theorem}

\usepackage{enumitem}
\usepackage{color}
\usepackage{titlesec}
\usepackage{xcolor}
\usepackage{wasysym}

\newtheorem{definition}{Definition}
\DeclareMathOperator*{\argmin}{arg\,min}

\usepackage[
backend=biber,
style=ieee,  
sorting=none         
]{biblatex}

\definecolor{myred}{RGB}{255,0,0}

\begin{document}

\title{Learning Unbiased Cluster Descriptors for Interpretable Imbalanced Concept Drift Detection}

\author{Yiqun~Zhang, \IEEEmembership{Senior Member,~IEEE}, Zhanpei~Huang, Mingjie~Zhao, Chuyao Zhang,\\ Yang~Lu, \IEEEmembership{Senior Member,~IEEE}, Yuzhu~Ji, \IEEEmembership{Member,~IEEE}, Fangqing~Gu,~and~An~Zeng, \IEEEmembership{Member,~IEEE}\\

\thanks{Received 20 January 2025; revised 24 June 2025; accepted 3 August 2025. This work was supported in part by the National Natural Science Foundation of China (NSFC) under Grants 62476063, 62302104, and 62376233, in part by the Natural Science Foundation of Guangdong Province under Grants 2025A1515011293 and 2023A1515012884, in part by the Natural Science Foundation of Fujian Province under Grant 2024J09001, in part by the Science and Technology Program of Guangzhou under Grant SL2023A04J01625, and in part by Xiaomi Young Talents Program. (Corresponding authors: Yuzhu Ji and An Zeng)

Yiqun Zhang, Zhanpei Huang, Mingjie Zhao, Chuyao Zhang, Yuzhu Ji, and An Zeng are with the School of Computer Science and Technology, Guangdong University of Technology, Guangzhou 510006, China. E-mail: yqzhang@gdut.edu.cn; \{2112405010, 2112205249, 2112305290\}@mail2.gdut.edu.cn; \{yuzhu.ji, zengan\}@gdut.edu.cn. 

Yiqun Zhang and Mingjie Zhao are also with the Department of Computer Science, Hong Kong Baptist University, Hong Kong SAR, China. 

Yang Lu is with the Key Laboratory of Multimedia Trusted Perception and Efficient Computing, Ministry of Education of China, Xiamen University, Xiamen 361000, China. E-mail: luyang@xmu.edu.cn. 

Fangqing Gu is with the School of Mathematics and Statistics, Guangdong University of Technology, Guangzhou 510006, China. E-mail: fqgu@gdut.edu.cn.

}}

\markboth{IEEE TRANSACTIONS ON EMERGING TOPICS IN COMPUTATIONAL INTELLIGENCE, AUGUST 2025}
{Shell \MakeLowercase{\textit{et al.}}: A Sample Article Using IEEEtran.cls for IEEE Journals}

\maketitle

\begin{abstract}
Unlabeled streaming data are usually collected to describe dynamic systems, where concept drift detection is a vital prerequisite to understanding the evolution of systems.
However, the drifting concepts are usually imbalanced in most real cases, which brings great challenges to drift detection. That is, the dominant statistics of large clusters can easily mask the drifting of small cluster distributions (also called small concepts), which is known as the `masking effect'.
Considering that most existing approaches only detect the overall existence of drift under the assumption of balanced concepts, two critical problems arise: 1) where the small concept is, and 2) how to detect its drift. 
To address the challenging concept drift detection for imbalanced data, we propose Imbalanced Cluster Descriptor-based Drift Detection (ICD3) approach that is unbiased to the imbalanced concepts. 
This approach first detects imbalanced concepts by employing a newly designed multi-distribution-granular search, which ensures that the distribution of both small and large concepts is effectively captured. Subsequently, it trains a One-Cluster Classifier (OCC) for each identified concept to carefully monitor their potential drifts in the upcoming data chunks.
Since the detection is independently performed for each concept, the dominance of large clusters is thus circumvented. ICD3 demonstrates highly interpretability by specifically locating the drifted concepts, and is robust to the changing of the imbalance ratio of concepts.
Comprehensive experiments with multi-aspect ablation studies conducted on various benchmark datasets demonstrate the superiority of ICD3 against the state-of-the-art counterparts.
\end{abstract}

\begin{IEEEkeywords}

Concept drift detection, imbalanced data learning, unsupervised learning, streaming data, cluster analysis.
\end{IEEEkeywords}

\section{Introduction}\label{section1_introduction}

\IEEEPARstart{U}NSUPERVISED streaming data analysis is a significant task to understand dynamic environment~\cite{lcd,ADSD,UDCL,UFS,SOHI}. With the environment evolving over time, the distributions of the corresponding dataset may keep changing correspondingly, which is known as concept drift~\cite{ACDWM,intro_23, ade,dirftlearn,zhao2023unsupervised}. When drift occurs, the use of the previously obtained data knowledge may hamper the effectiveness of data analysis~\cite{review2024,MDStream,CDDL}, and thus detecting the occurrence of concept drift becomes a vital premise. Due to the unavailability of labels, the unsupervised drift detection should be performed with concept exploration and drift detection, separately on the data samples collected within adjacent time frames, which is also called data chunks~\cite{reviewKBS2022, CDTD}. 
The uncertainty introduced by the unsupervised concept exploration phase makes drift detection more challenging than in supervised scenarios.

\begin{figure}[!t]
\centering
\includegraphics[width=3.5in]{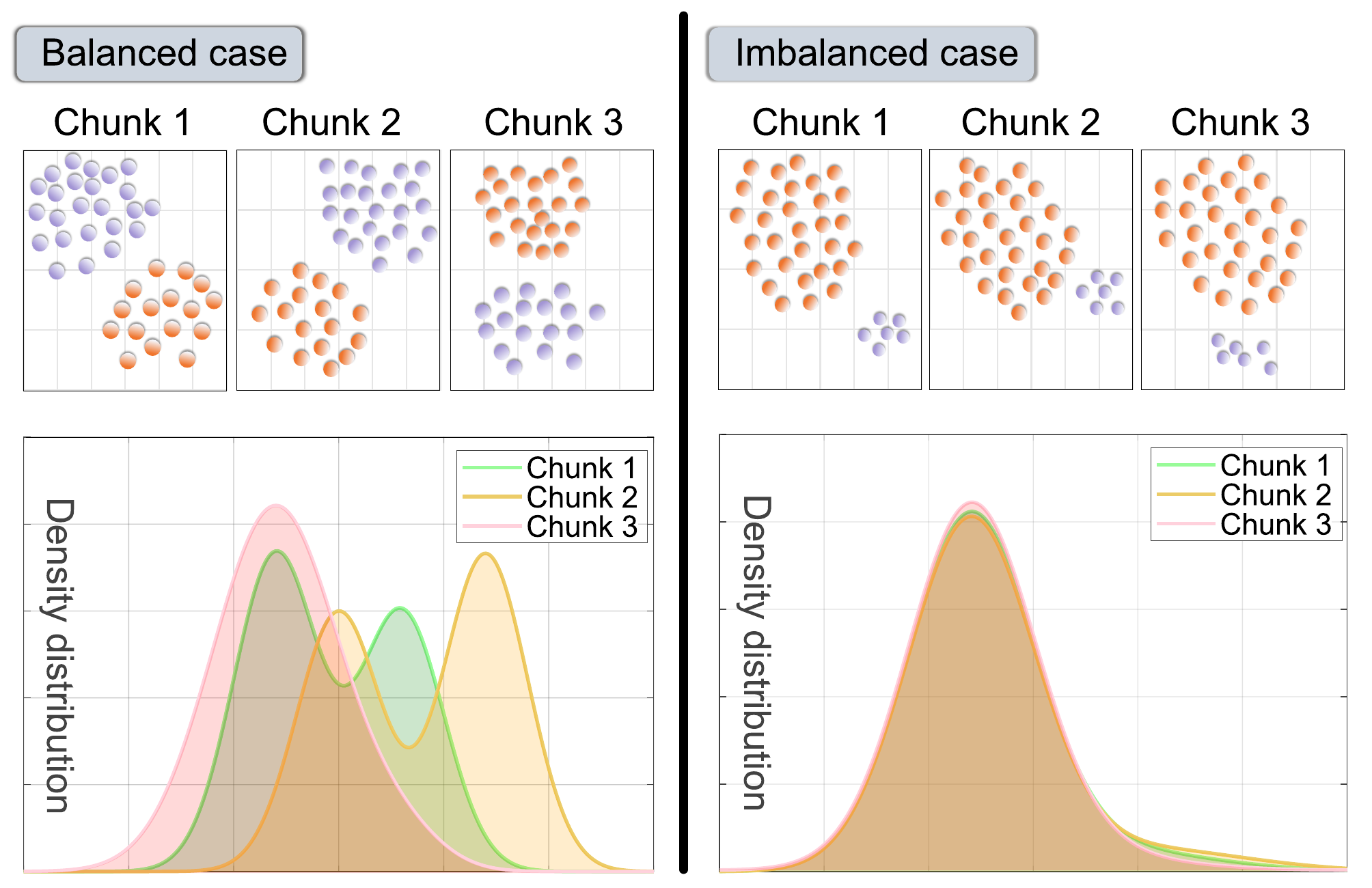}
\caption{Comparison of drift detection in balanced and imbalanced cases. The chunk-wise drift detection can easily become incompetent under the imbalanced case because even though a significant drift appears on a small cluster, the corresponding impact can almost be masked by the large clusters in terms of the overall density distribution.}
\label{fig_toyexample_AR}
\end{figure}

Existing unsupervised drift detection approaches can be roughly categorized into model-based~\cite{MD3,D3,OCDD,MMBCD} and statistical test-based~\cite{kdqtree,QuanTree,EI-Kmeans,QT-EWKM} detection approaches, according to their adopted way of concept description. The former type models the chunk distribution and then monitors the difference in terms of model outputs between two data chunks, while the latter stream statistically describes the distributions of chunks and tests to validate the null hypothesis of drift occurrence. However, the majority of the existing approaches tend to detect the overall chunk-wise concept changes, thus overlooking the exact locations of drifted distribution regions. Consequently, these approaches exhibit high susceptibility to imbalanced concept distributions, as fluctuations in small clusters minimally impact the global distribution.

Specifically, model-based methods detect drift by monitoring prediction errors across entire incoming data chunks using models trained on previous chunk distributions~\cite{oudd}. A larger prediction error rate in the new chunk surely indicates the overall distribution inconsistency between two adjacent chunks. Among this type of methods, MDDD~\cite{MD3}, and OCDD~\cite{OCDD} adopt a common basic idea to detect the drift of an incoming chunk according to a classification model, which is trained on the previous chunk. If the classification error rate of the samples from the incoming chunk reaches a threshold, a drift is considered to be detected. A slightly different strategy is adopted by the method that trains a binary drift detection discriminative classifier~\cite{D3}. It classifies the data samples from the previous and the new incoming chunks, and recognizes drift occurrence when the two chunks can be easily separated. However, all these methods use a single model to overall describe the distributions of entire chunks, which can easily overlook the distribution changes that appear on relatively small clusters due to their relatively low contribution to the overall prediction error.

By contrast, statistical test-based approaches perform drift detection by directly comparing the distributions corresponding to two data chunks and judging the significance of the drift through statistical tests. For instance, the conventional MWW~\cite{MWW} and FKS~\cite{FKS} directly employ the Mann-Whitney-Wilcoxon test~\cite{MWW_test} and Kolmogorov-Smirnov test~\cite{KS_test}, respectively, to assess the inconsistency of two data chunks in terms of their distributions. Recent approaches including QT-EWMA~\cite{QT-EWKM}, EI-KMeans~\cite{EI-Kmeans}, QuantTree~\cite{QuanTree}, and KdqTree~\cite{kdqtree}, further construct histograms to more specifically describe the density of sub-distributions within data chunks, and then perform significance test to know how different two chunk distributions are. Although the advanced test-based approaches describe the local data distributions, the description strategy, e.g., grid partition of the distribution space, is uniformly performed, and thus incompetent in reflecting the changes in small data sample clusters. 

All the above-mentioned advances perform concept drift detection under an implicit assumption that the sub-distributions within a data chunk are roughly balanced~\cite{CDDMI} as shown in the left panel of Fig.~\ref{fig_toyexample_AR}. Furthermore, the mentioned current methods all determine whether drift occurs only, without understanding the drifts about where the drift arises and what the new concept distribution looks like. Therefore, they are neither competent to the drift detection of imbalanced clusters nor can they locate where the drift occurs. This greatly limits their application in real scenarios where imbalanced cluster drifts are very common~\cite{lid,ACICD,ICILDS}, e.g., a drifting small cluster corresponding to a small number of COVID-19 patients with continuously evolving virus strain next to the large cluster of healthy people. 
It can be seen from the right panel of Fig.~\ref{fig_toyexample_AR} 
that although the small cluster undergoes significant drift, the overall density distribution remains almost unchanged due to the 'masking effect' caused by the relatively large cluster. This highlights the difficulty of detecting drifts occurring in relatively small concepts, which are often overlooked in unsupervised concept drift detection tasks.

This paper, therefore, introduces the Imbalanced Cluster Descriptor-based Drift Detection (ICD3) approach for robust and interpretable drift detection of imbalanced data. ICD3 first implements the proposed Density-guided Concept Distribution Learning (DCDL) algorithm that finely partitions the dataset into many small clusters and then hierarchically merges the densely connected small clusters to simultaneously ensure the exploration of small clusters and avoid unreasonably dividing large clusters. Then the distribution changes are independently tracked within each obtained cluster by training a One-Cluster Classifier (OCC) and monitoring its prediction accuracy. The OCCs act to regularize the contributions of imbalanced concepts into the same level, and thus the `masking effect' brought by the larger clusters on the smaller ones can be adequately relieved. More importantly, since ICD3 separately monitors the drifts in each concept, it can precisely answer the questions of whether and where the drift occurs, and how the drifted region looks like. By working in a detect-then-train manner for streaming data chunks, ICD3 can keep robustness in drift detection and description. Extensive experiments including performance comparison with state-of-the-art counterparts, ablation studies, parameter sensitivity evaluation, and qualitative results visualization are conducted on 14 benchmark datasets to demonstrate the superiority of our approach. Moreover, an imbalanced concept drift generator is developed to mimic extremely imbalanced concepts, which can serve as a general experiment tool for subsequent drift detection research. The main contributions can be summarized into four-fold:

\begin{itemize}

  \item \textbf{A new concept drift detection paradigm:} Compared with the existing discriminative drift detection methods, we propose a new generative paradigm, which first describes each of the imbalanced concepts and then tracks them for accurate and explainable drift detection.

  \item \textbf{Unbiased drift detection:} A multi-granular concept detection strategy has been proposed. It learns excessive prototypes to represent the potential small concepts, and also merges them to describe larger concepts with arbitrary shapes, facilitating unbiased concept drift detection within the incoming chunk.

  \item \textbf{Interpretable drift monitoring:} Based on the learned cluster descriptors, an interpretable drift monitoring mechanism has been designed. Despite its timely drift alarm ability, it can also precisely locate where the drift occurs and intuitively visualize the drifted regions.
  \item \textbf{Robust to drift types:} By comparing how incoming samples deviate from the corresponding cluster descriptors learned on the base chunk, the proposed method can precisely detect the occurrence of tiny drift, thus robust to sudden, gradual, incremental, and recurring concept drift.

\end{itemize}

The remainder of this paper is organized as follows. Section II overviews the existing related work. Section III presents the details of the proposed ICD3 method. The experimental results are demonstrated and discussed in section IV. Finally, we conclude in Section V.

\section{Related Work}

This section provides an overview of the related works on concept drift detection and imbalanced cluster analysis.

\subsection{Concept Drift Detection}

Based on the indicators for identifying concept drift, existing concept drift detection methods can be roughly categorized into two types: model-based and test-based. The model-based methods monitor the performance changes in terms of the model prediction.
The method proposed in~\cite{DDM} tracks the error rate and reports the occurrence of a drift when the error rate exceeds a pre-set threshold. To improve the performance in detecting gradual drifts, the work in~\cite{EDDM} further monitors the distribution of correctly classified samples to detect gradual drift, while also introducing the sensitivity to noise.
Accordingly, an adaptive windowing method~\cite{ADWIN} circumvents the noise issue by comparing the prediction error rate between adjacent windows. Despite the effectiveness of the above-mentioned methods, they usually require data labels, which limits their practicality. 

To address this limitation, the method~\cite{D3} employs a binary classifier trained on the data samples from two adjacent temporal windows, and reports a drift if the data samples from the two windows can be well separated. Meanwhile, One-Class Concept Drift Detection (OCDD) proposed in~\cite{OCDD} adopts One-Class Classification paradigm to flag drifted samples as anomalies, and reports a drift when the anomaly scale surpasses a certain threshold. ClusterSVDD~\cite{occ-cluster} extends this idea by modeling multiple data modes with hyperspheres to enhance drift detection robustness. More recently, representation-based methods such as MCD~\cite{mcd} leverage an ensemble of autoencoders to encode latent data distributions, offering high flexibility in adapting to diverse and complex drift patterns. However, it exhibits limited interpretability and sensitivity, as the focus on dominant patterns within the latent space may cause subtle drifts to be masked.

The test-based methods focus on the statistical significance of the occurrence of concept drifts. Approaches proposed in \cite{FKS} and~\cite{MWW} adopt Kolmogorov-Smirnov test and Mann-Whitney-Wilcoxon test, respectively, to determine concept drift. However, they can only indicate distribution differences but are incompetent in providing explainable detailed information about the distribution changes. 
By contrast, the work proposed in~\cite{kdqtree} partitions data into different local sub-regions using KdqTree formed by combining the structures of kd-tree and quad-tree, and creates a density distribution histogram to monitor drift through hypothesis testing. Similarly, the Qtree~\cite{QuanTree} and QT-EWMA~\cite{QT-EWKM} approaches use quad-tree-based partition to grasp more detailed distribution information for the following tests. However, when encountering relatively complex distributions, tree-based approaches will produce partitions that are not intuitively relevant to the exact concept distributions. They could also cause drift blind spots at the leaf nodes, which thus hamper their interpretability and accuracy, respectively. To tackle these problems, Equal Intensity $k$-Means (EI-KMeans)~\cite{EI-Kmeans} further adopts density-weighted $k$-means algorithm and Pearson's chi-squared test to more appropriately characterize drifts. Nevertheless, the amplify partition strategy it adopts could split one compact concept into multiple sub-clusters, and thus introduce randomness into the interpretability and accuracy of its detection results. 

To sum up, all the above-mentioned two types of methods are designed for global drift detection and overlook the imbalance issue, leaving considerable improvement room in terms of the interpretability and accuracy of concept drift detection. In particular, these methods fail to effectively detect drift when it occurs within minority concepts. As a result, delayed or missed drift detection may frequently occur in such case.

\subsection{Imbalanced Cluster Analysis}

Imbalanced data distributions are common in real-world data analysis tasks. Under unsupervised scenarios, existing clustering methods like $k$-means tend to exhibit a bias toward the exploration of large clusters (so-called `uniform effect')~\cite{xiongCV}. To address this issue, fuzzy $k$-means~\cite{liang2012} has been proposed featuring relatively discriminative in detecting imbalanced clusters. However, since all the $k$-means type algorithms have bias to the convex spherical-shaped clusters, the distribution density-guided clustering algorithms, i.e., affinity propagation~\cite{apcluster} and multi-exemplar merging clustering~\cite{memc,zhang2025asynchronous,zhang2025adaptive,cai2024robust,cheung2018fast} have been developed in the literature. They adopt a common basic idea to first group data samples into sub-clusters, and then merge the densely connected sub-clusters according to their density distributions. 
More recently, a method called SMCL~\cite{SMCL} has been proposed based on competitive learning theory~\cite{RPPCL}. It initially groups imbalanced data into several compact sub-clusters represented by a series of self-organizing prototypes trained through competitive clustering algorithm. Following this, it merges the sub-clusters to form final clusters, and also determines the optimal number of clusters based on the separation degree among the clusters. However, all the above-mentioned approaches are designed for the cluster analysis of static data only, and cannot be simply used for the concept drift detection of streaming data.

\begin{figure*}[t]
\centering
\includegraphics[width=6in]{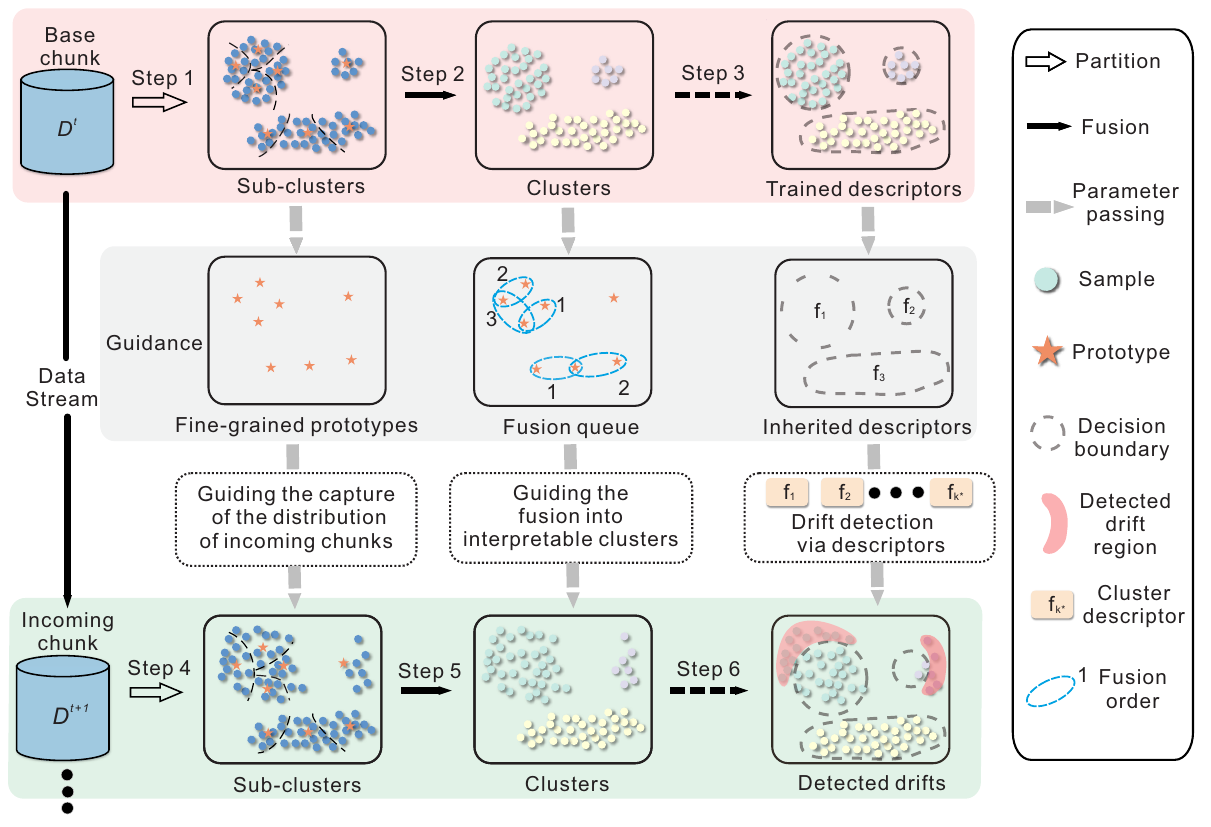}
\caption{Overall workflow of ICD3. ICD3 works on two adjacent chunks. In the base chunk $D^t$, fine-grained prototypes are first learned and used to partition the samples into multiple sub-clusters. Next, fusion queues are learned to merge these sub-clusters into interpretable clusters. Following this, descriptors are trained for each of the clusters. The learned prototypes guide the capture of the distribution in the incoming chunk $D^{t+1}$, while the fusion queues guide the merging of sub-clusters into clusters. Concept drift detection is performed through the descriptors inherited from the base chunk.}
\label{fig_workflow}
\end{figure*}

\section{Propose Method}

\begin{table}[t]
\caption{Frequently used notations and symbols.}
\label{tab_sample_explain}
\centering
\begin{tabular}{l|l}
\toprule
Notations and Symbols &Explanations\\
\midrule
Subscript e.g.``$h$'' of $\mathbf{x}_h$ &Element index indicator\\
Superscript, e.g.``$t$'' of $D^t$ &Chunk index indicator\\
Angle-bracketed superscript, e.g.``$*$'' of $k^*$ &Annotation\\
Parentheses, e.g. $F_i(\cdot)$, $L_i(\cdot)$ &Function\\
Greek letters, e.g. $\alpha$, $\beta$, and $\theta$& Coefficients \\
Uppercase, calligraphic font, e.g. $\cal Q$ & Queue set\\
Uppercase, e.g. $D, C,$ and $G$ &Set\\
Lowercase, e.g.  $n$, $k$ &Value\\
Lowercase, bold font, e.g. $\mathbf{x}_h$ &Vector\\
\bottomrule
\end{tabular}
\end{table}

The problem of imbalanced concept drift detection is first formulated in Section~\ref{section3_Preliminary}. Then Sections~\ref{section3_imbalance_division} and~\ref{section3_descriptor_learning} provide the technical details of concept distribution learning and concept descriptor training, respectively. Section~\ref{section3_MICD_drift_detection} describes the concept drift detection and understanding processes with time complexity analysis. Here we first provide an overview of the proposed ICD3 method in Fig.~\ref{fig_workflow}. With streaming data chunks flowing in, ICD3 works in a detect-then-train manner. That is, if drift is detected in the current chunk, then treat it as a base chunk to update the concept description for subsequent drift detection. Further analysis can be conducted to exactly locate and visualize the detected drifts.

\subsection{Problem Definition}\label{section3_Preliminary}

Given a streaming dataset $D=\{D^1, D^2,\ldots, D^t,\ldots\}$ composed of samples collected in a chunk-by-chunk manner, where $D^t$ is a set of $n^t$ data samples collected at $t$-th time slot. We denote $h$-th data sample in $D^t$ as a feature vector $\mathbf{x}_h$. Within a chunk, samples can be partitioned into $k$ clusters denoted as $C = \{ C_1, C_2, \ldots, C_k \}$. Concept drift refers to the phenomenon where the data distribution becomes inconsistent over adjacent chunks. More specifically, when the joint probability satisfies $p_{t}({\mathbf{x}},C) \ne p_{t-1}({\mathbf{x}},C)$, it indicates that concept drift occurs in $D^t$ w.r.t. $D^{t-1}$. With consecutive drifting, clusters can further become severely imbalanced in terms of the cluster size, which can be expressed as $p({C}_i|D^t) \ll p({C}_j|D^t)$. 

Given two temporarily adjacent chunks, we uniformly denote the previous one as a base chunk $D^b$ and the incoming chunk as $D^m$ for simplicity without loss of generality. Then the concept drift detection problem that will be solved can be summarized as how to answer the following questions:

Q1: \textbf{What the concepts are?} A specific cluster can represent a particular concept, and the general goal of clustering is to minimize the distance between samples and their cluster centers, formalized as:
\begin{equation}
    \min \sum_{i=1}^k \sum_{\mathbf{x}_h \in  C_{i}} \| \mathbf{x}_h - C_{i} \|^2.
\end{equation}
However, when $p(C_i|D^t) \ll p(C_j|D^t)$ occurs, indicating an imbalance, conventional methods struggle to accurately represent a concept, necessitating a clustering approach capable of unbiasedly representing such imbalanced clusters. 

Q2: \textbf{Whether drift occurs?} When incoming chunk $D^m$ arrives, it is crucial to detect whether concept drift occurs. Particularly under imbalanced conditions, it is essential to ensure that drift can be identified promptly.

Q3: \textbf{Where does the drift occur?} Furthermore, detecting concept drift requires identifying which specific cluster $C_i^m$ within the incoming chunk $D^m$ experiences drift.

Q4: \textbf{How does the drifted region looks like?} Following the detection of concept drift in cluster $C_i^m$, it is of necessity to further identify drift samples:
\begin{equation}
     \mathcal{D}^m_i = C_i^m \setminus (C_i^m \cap C_i^b)
\end{equation}
to precisely characterize drifted region.

To address these challenges, the approach involves three main steps: 1) clustering algorithm is applied to the base chunk $D^b$ to represent concepts of varying sizes (detailed in Section \ref{section3_imbalance_division}), 2) a descriptor is created to track each concept (detailed in Section \ref{section3_descriptor_learning}), and 3) drift detection is performed on incoming chunk $D^m$. Once the drift is identified, specific clusters experiencing drift are located, and within these clusters, subsets of drifted samples are identified to delineate the drifted regions (detailed in Section \ref{section3_MICD_drift_detection}).

\subsection{DCDL: Density-Guided Concept Distribution Learning}\label{section3_imbalance_division}

To learn the imbalanced and drifting concept distributions adaptively and flexibly, we adopt a competitive-penalize learning mechanism that initializes amplified prototypes to describe the data distribution and then lets the prototypes automatically compete to eliminate redundant ones. However, the existing competitive-penalize learning has common drawbacks w.r.t. the problem to be solved: 1) \textbf{Sensitive to the initialization:} even the existing advanced initialization methods~\cite{in_kmeans2,in_kmeans3,in_kmeans4,in_kmeans5} exhibit bias, tending to assign prototypes to larger clusters, which hinders the exploration of imbalanced clusters; 2) \textbf{Granularity restriction:} the scale of the initialized prototypes constrain the granularity of explored clusters, thus preventing the detection of micro-clusters; 3) \textbf{Uniform effect:} as a k-means-type algorithm, it inherits the uniform effect~\cite{xiongCV}, i.e., implicitly assuming that clusters are balanced in size, thus failing to effectively detect imbalanced clusters.

Accordingly, we propose a density measure to guide competitive-penalize learning, as local density peaks can effectively reflect cluster centers without a bias towards the large clusters. Subsequently, to avoid granularity limitation, we utilize an incremental strategy that identifies the distribution of concepts by increasing the number of prototypes, allowing for a more comprehensive exploration of clusters. Finally, we implement a fusion strategy that enables prototypes to be progressively merged, ensuring the representation of concepts with varying sizes.

To identify samples with higher local density~\cite{DPC_clustering}, the reverse nearest neighbors ($\mathcal{RNN}$) is first defined.

\begin{definition}[Reverse Nearest Neighbors]
    \label{definition_RN}
    Given a data chunk $D^b=\{\mathbf{x}_1,..., \mathbf{x}_{n^b}\}$, $d_{h{\mathfrak{h}}}$ represents the Euclidean distance between $\mathbf{x}_h$ and $\mathbf{x}_{\mathfrak{h}}$. The reverse nearest neighbors of sample $\mathbf{x}_h$ are defined as a sub-set $\mathcal{RNN}(\mathbf{x}_h)$, formed as:
   \begin{equation} \label{eq_rnn}
    \mathcal{RNN}(\mathbf{x}_h)   = 
   \{ \mathbf{x}_{\mathfrak{h}} \in D^b \setminus \{\mathbf{x}_h\} \mid \mathbf{x}_h = \argmin\limits_{\mathbf{x}_{\mathfrak{l}} \in D^b \setminus \{\mathbf{x}_{\mathfrak{h}}\}} d_{{\mathfrak{h}}{\mathfrak{l}}}\}. 
    \end{equation} 
    Here, a sample $\mathbf{x}_{\mathfrak{h}}$ is included in the reverse nearest neighbors $\mathcal{RNN}(\mathbf{x}_h)$ only if $\mathbf{x}_h$ is the closet sample to $\mathbf{x}_{\mathfrak{h}}$.
\end{definition}

 Utilizing sub-set $\mathcal{RNN}$ to define local density is parameter-free, grounding density estimation directly in neighbor relationships. For a sample $\mathbf{x}_h$, the samples within $\mathcal{RNN}(\mathbf{x}_h)$ provide a more representative understanding of $\mathbf{x}_h$ by incorporating those that identify $\mathbf{x}_h$ as the closest neighbor, thus enhancing the emphasis on global structural information. 
 
After obtaining the reverse nearest neighbors for each sample, the local density $\rho$ can be formulated as:
\begin{equation}
    {\rho}_h = \exp \left( { -  {\frac{1}{{\left| {\mathcal{RNN}\left( {{{\mathbf{x}}_h}} \right)} \right|}}\sum\limits_{{{{\mathbf{x}}_{\mathfrak{h}}}} \in \mathcal{RNN}\left( {{{\mathbf{x}}_h}} \right)}} {{d_{h{\mathfrak{h}}}}} }  \right),
    \label{eq_compute_loc_dens}
\end{equation}
where $\left| {\rm{\cdot}} \right|$ denotes the number of samples in ${\mathcal{RNN}\left( {{{\mathbf{x}}_h}} \right)}$. For cases where there are no samples in the reverse nearest neighbors (i.e., ${\left| {\mathcal{RNN}\left( {{{\mathbf{x}}_h}} \right)} \right|}=0$), the local density is set to 0. The higher local density ensures that the prototype represents a local prominent sample distribution, while the separation between it and the other prototypes relieves the case that a large compact cluster is over-represented by multiple prototypes. This mechanism enables the identification of high local density samples, even for clusters with fewer samples.

Subsequently, the density gap $\delta$ can be computed by:
\begin{equation}
    {\delta_h} = \mathop {\min }\limits_{{\mathfrak{h}}: \rho_{\mathfrak{h}} > \rho_h}({{d_{h{\mathfrak{h}}}})} ,
    \label{eq_compute_dens_gap}
\end{equation}
where $\delta_h$ is the minimum distance from $\mathbf{x}_h$ to any sample $\mathbf{x}_{\mathfrak{h}}$ with higher local density $\rho_{\mathfrak{h}}$.  
For samples with the highest local density, the density gap is defined as the maximum distance to any other sample $\mathbf{x}_{\mathfrak{h}}$, expressed as $\delta_h = {\max_{\mathfrak{h}}}\left( {{d_{h{\mathfrak{h}}}}} \right)$. Accordingly, the top $\kappa_0$ samples with the highest density gaps are selected as the initial prototypes, denoted as  ${S = \{ }{{\mathbf{s}}_1}{\rm{,}}...{\rm{,}} {{\mathbf{s}}_{\kappa_0}}{\} }$. Such density-guided prototype initialization is insensitive to cluster sizes, ensuring that small clusters are also adequately represented by the prototypes.

\begin{algorithm}[!t]
\SetAlgoLined 
\SetKwInOut{Input}{Input}\SetKwInOut{Output}{Output}
 \caption{\selectfont{Concept Distribution Learning}} \label{alg_DCDL}
\Input{Base chunk $D^b=\{\mathbf{x}_1 ,..., \mathbf{x}_{n^b}\}$, number of initial prototypes $\kappa_0$, maximum number of iterations $l$, reverse nearest neighbors $\mathcal{RNN}$ for samples in $D^b$.}
\Output{Optimal number $k^*$, fusion queue ${\mathcal{Q}^b}$, Prototype list $S$, and Clusters $G^b$.}

Initialize winning time $w \leftarrow 0$;\\
\For{$h \leftarrow 1$ \KwTo $n^b$}{
    Compute $\delta_h$ for $\mathbf{x}_h$ according to Eq.~(\ref{eq_compute_dens_gap});\\
}
Select the top $\kappa_0$ samples with the largest density gaps $\delta_h$ as initial prototypes $S$;\\

\For{$\tau \leftarrow 1$ \KwTo $\mathcal{I}$}{
    \For{$h \leftarrow 1$ \KwTo $n^b$}{
        Update $S$ and $w$ according to Eqs.~(\ref{eq_Choose_seed_Point}) and (\ref{eq_self-adaptive_penalized_competitive_learning});\\
    
    }
    \eIf{$\exists \ w=0$}{Break;}{
        Add a new prototype $s'$ and update $S = S \cup \mathbf{s}'$;\\
}

}
Assign samples in $D^b$ into clusters in $C^b$ according to $S$ by Eq.~(\ref{eq_Choose_seed_Point});\\
\For{$\varepsilon \leftarrow k$ \KwTo $2$}{
    Select $G_i^b$, $G_j^b$ with minimum separation according to Eq.~(\ref{alg_sep});\\
    $G^b \leftarrow G^b \setminus \{G_i^b, G_j^b\} \cup \{G_i^b \cup G_j^b\}$;\\
}

Select $k^*$ according to Eq.~(\ref{eq_compute_K}), obtain $G^b$ and ${\mathcal{Q}^b}$;\\
\Return{$k^*$, ${\mathcal{Q}^b}$, $S$, $G^b$};

\end{algorithm}

With the density-guided prototypes $S$, we further incrementally amplify more prototypes accordingly to avoid the granularity restriction. The adjustment of prototype positions is based on the relationship between samples and prototypes: winning prototypes are moved closer to their corresponding samples while the others are pushed further away. Specifically, given $S$, each sample ${\mathbf{x}}_h$ in $D^b$ is assigned to a winning prototype ${\mathbf{s}}_j$ according to the following rule:
\begin{equation}
    \label{eq_Choose_seed_Point}
    {r_{j,h}} = \left\{
    {\begin{array}{lll}
        1,&&{\text{if }j = argmi{{\rm{n}}_{1 \le \mathfrak{d} \le \kappa_0}}(||{{\mathbf{s}}_\mathfrak{d}} - {{\mathbf{x}}_h}|{|^2})}\\
      0,&&\text{otherwise}
    \end{array}},
\right. 
\end{equation}
where $\mathbf{x}_h$ is the $h$-th sample in $D^b$, $\mathbf{s}_j$ is the $j$-th prototype. $r_{j,h}=1$ means that $ \mathbf{s}_j$ is the winning prototype and ${{\mathbf{x}}_h}$ is assigned to $ \mathbf{s}_j$. If $r_{j,h}=0$, it indicates the opposite. Each prototype records a winning time $w$, which is incremented by 1 once each sample is assigned to it, quantifying the prototype's effectiveness. The adjustment process for the prototypes can be expressed as:
\begin{equation}
    \label{eq_self-adaptive_penalized_competitive_learning}
    \mathbf{s}_j^{\text{new}} = \left\{
    {\begin{array}{lll}
        \mathbf{s}_j^{\text{old}} + \alpha (\mathbf{x}_h - \mathbf{s}_j^{\text{old}}),&\text{if} \ r_{j,h}=1 \\
        \mathbf{s}_j^{\text{old}} - \beta_j \alpha (\mathbf{x}_h - \mathbf{s}_j^{\text{old}}),&\text{otherwise}
    \end{array}}.
\right. 
\end{equation}
For each sample $\mathbf{x}_h$ within $D^b$, $S$ is updated based on Eq.~(\ref{eq_self-adaptive_penalized_competitive_learning}). Here, $\alpha$ is the learning rate of the prototypes, and $\beta_j$ is a dynamic competitive penalty coefficient calculated as:
\begin{equation}
    \beta_j = \exp\left(-\frac{\|\mathbf{s}_j - \mathbf{x}_h\|^2 - \|\mathbf{s}_v - \mathbf{x}_h\|^2}{\|\mathbf{s}_v - \mathbf{s}_j\|^2}\right)
    \label{eq_beta_j},
\end{equation}
where $\mathbf{s}_v$ indices the winning prototype of $\mathbf{x}_h$. Samples positioned closer to the midpoint between $\mathbf{s}_j$ and $\mathbf{s}_v$ contribute to a larger $\beta_j$, thereby imposing a stronger competitive penalty. In the subsequent cases, when samples are situated between two prototypes within the same cluster, they exhibit higher density, thereby incurring greater intra-cluster competitive penalties. Conversely, when samples are positioned between prototypes belonging to different clusters, they are relatively sparse, leading to reduced inter-cluster competitive penalties.

In each round of competition, prototypes with winning time $w=0$ indicates that they have been eliminated in the competition, and the other prototypes with $w\neq0$ are sufficient to represent the concepts. Conversely, if all prototypes have $w>0$, it suggests that the current number of prototypes is insufficient to adequately capture the local concepts. Accordingly, more prototypes should be added for competitive learning to facilitate a finer-grained cluster description. With the above process, the fine granularity prototypes ${S = \{ }{{\mathbf{s}}_1}{\rm{,}}...{\rm{,}} {{\mathbf{s}}_\kappa}{\rm{\} }}$ are obtained, and the samples in $D^b$ are assigned to the $\kappa$ prototypes, forming $k$ sub-clusters ${C^b} = \{ C_1^b,..., C_k^b\} $.

Density-guided initialization ensures unbiased prototype selection by leveraging local density peaks. However, static granularity fails to represent evolving and complex concepts. To address this limitation, incremental competitive-penalize learning is employed, enabling prototypes to dynamically adjust their positions and incrementally increase in quantity through competitive adjustments, thereby representing concepts more appropriately.

Incremental competitive learning is utilized to obtain fine-grained prototypes that appropriately represent concepts. However, competitive learning, inherits the uniform effect of k-means, hindering the detection of imbalanced clusters. After obtaining fine-grained partitioning results, a fusion strategy is used to combine selected prototypes, forming better representations for larger concepts.

Following this, we analyze the separation $\zeta_{ij}$ between pairs of sub-clusters. Specifically, the prototypes $\mathbf{s}_i$ and $\mathbf{s}_{j}$ are projected to the position $ - 0.5$ and $0.5$, respectively, while the samples belonging to the sub-clusters $C_{i}^b$ and $C_j^b$ are projected onto the line segment connecting $\mathbf{s}_i$ and $\mathbf{s}_j$. These projections are utilized to estimate the discrete probability density function $\phi(\cdot)$ based on a binary Gaussian mixture model. The separation $\zeta_{ij}$ are calculated as:
\begin{equation}
\label{alg_sep}
    \zeta_{ij}=\min \frac{1}{\phi(A)},
\end{equation}
where $A$ is an interval with step $0.01$ from $ - 0.5$ to $0.5$. 

Subsequently, sub-clusters with minimum separation are iteratively selected for prioritized fusing until only one cluster remains. After each fusion step, the separability and compactness of the overall clustering structure are evaluated~\cite{SMCL}. The optimal number of clusters $k^*$ is selected according to:
\begin{equation}
    {k^*} = \mathop {\argmin }\limits_{1 \le \varepsilon \le k - 1} \left( \frac{{se{p_\varepsilon}}}{{{{\max }_{\varepsilon'}}\{ se{p_{\varepsilon'}}\} }} + \frac{{co{m_\varepsilon}}}{{{{\max }_{\varepsilon'}}\{ co{m_{\varepsilon'}}\} }}\right)
    \label{eq_compute_K},
\end{equation}
where $sep_\varepsilon$ and $com_\varepsilon$ quantify the overall separability and compactness, respectively, for a clustering result with $\varepsilon$ prototypes. As for the denominators, they are employed to normalize the values to a common range by dividing each molecule by the corresponding maximum value. The final clustering results with an optimal number (i.e., $k^*$) of clusters have been formed for appropriately representing the cluster distribution of $D^b$, which is denoted as ${{G}^{b}} = \{ G_1^b,..., G_{{k^*}}^b\}$. 

During the merging process, a set of fusion queues ${{\mathcal{Q}}^{b}} = \left\{ {{\mathbf{q}_1^b},...,{\mathbf{q}_{{k^*}}^b}} \right\}$ is recorded corresponding to each of the final $k^*$ clusters, and each queue $\mathbf{q}_i^b$ records the sequential number of prototypes that are merged to form $G_i^b$. In subsequent concept drift detection (detailed in Section~\ref{section3_MICD_drift_detection}), the queues ${{\mathcal{Q}}^{b}}$ and prototypes $S$ are utilized in the incoming chunk $D^m$ to guide the formation of cluster concepts for drift detection. The whole process of the density-guided concept distribution learning in this section is summarized in Algorithm~\ref{alg_DCDL}.

So far, we have learned to describe imbalanced concepts, bringing clusters of varying sizes to an equal standing. This establishes a solid foundation for unbiasedly training independent descriptors for each cluster and enables effective drift detection in the subsequent processes.

\subsection{OCCL: One-Cluster Classifier Learning}\label{section3_descriptor_learning}
   
To finely describe each cluster concept within a chunk for accurate and interpretable drift detection, each cluster in $G^b$ obtained in Section~\ref{section3_imbalance_division} is trained with an OCC. For cluster $G_i^b$, all samples within it are treated as positive samples to train the corresponding OCC $F_i$ with objective: 
\begin{equation}\label{eq_occ}
\min L_i(\mathbf{x}) + \lambda \sum_{\omega=1}^{n^b} \xi_\omega, 
\end{equation}
where $L_i(x)$ is designed to find the optimal decision boundary of $G_i^b$, $\xi_\omega $ is the $\omega$-th slack variable that allows for flexibility in the model, and $\lambda$ is the trade-off parameter. 
Any standard classifier can be adapted into a one-cluster classifier to identify the distribution of a specific cluster, allowing for the flexible selection of classifiers for different scenarios.

Accordingly, a series of trained OCCs $F=\{F_1,F_2,...,F_{k^*}\}$ are obtained, which are also called single concept/cluster descriptors interchangeably. They provide a compact and conservative representation of the concepts within the base chunk. Subsequently, they are utilized to determine whether a newly incoming chunk exhibits drift.

\begin{figure}[!t]
\centering
\includegraphics[width=3.5in]{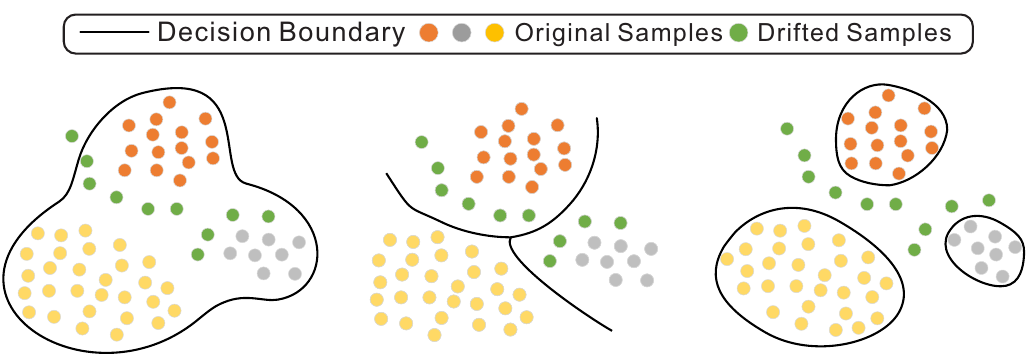}
\caption{Comparison of: 1) one OCC for one chunk, 2) multi-class classifiers, and 3) one OCC for one cluster, in the scenario of concept drift detection. }
\label{fig_M_o}
\end{figure}

\begin{remark}[Advantages of multiple OCCs in concept drift detection.]
    As shown in Fig.~\ref{fig_M_o}, OCCs provide a more flexible and fine-grained solution for handling concept drift. The single One-Cluster Classifier (left) is limited to detecting overall anomalies and struggles to distinguish changes in specific clusters. The multi-class classifier (middle), while capable of modeling all clusters jointly, is less adaptable to changes such as shifting cluster boundaries in dynamic environments. In contrast, the proposed OCCs (right) independently model each cluster, reducing interference between clusters and enabling more precise detection of specific concept changes. 
\end{remark}

\subsection{Concept Drift Detection and Positioning}\label{section3_MICD_drift_detection}

This section focuses on detecting concept drift in the new incoming chunk according to OCCs trained for the previous base chunk. In general, the newly incoming data samples should be partitioned directly to the $k^*$ optimal clusters $G^b$, and subsequently utilize OCCs for predictions. It is noteworthy that this strategy is straightforward and effective for convex, and spherical clusters. However, for non-convex and density-connected clusters, the cluster formation depends not only on distance but also on the density structure. Simply assigning new samples based on the Euclidean distance to a single cluster can lead to notable biased classification, consequently rendering the one-cluster classifiers ineffective.

Therefore, we utilize the fine-grained prototypes $S$ obtained from section~\ref{section3_imbalance_division} to partition the samples in the incoming chunk $D^m$ into $k$ sub-clusters using Eq.~(\ref{eq_Choose_seed_Point}), denoted as ${C^m} = \{ C_1^m,...,C_k^m\}$. Then we apply the prototype merging described in section~\ref{section3_imbalance_division}, i.e., using the merging queues in ${{\cal Q}^{b}}$, to merge the sub-clusters in ${C^m}$ as:

\begin{equation}
    \label{eq_merage_subclusters}
    {u_{i,j}} = \left\{ {\begin{array}{lll}
1,&{{\text{ if }}j \ { \in } \ {\mathbf{q}_i^b}}\\
0,&{{\text{ otherwise}}}
\end{array}} \right.,
\end{equation}
where $i=\{1,\cdots,k^*\}$, $j=\{1,\cdots,k\}$, and $u_{ij}=1$ indicates that $C_j^m$ should be fused into $G_i^m$, otherwise not.

\begin{remark}[Unbiased sample partition in $D^m$]
   The strategy of employing fine-grained prototype partitioning followed by merging embodies an unbiased cluster concept description for $D^m$. This approach captures various concepts within $D^m$ and reduces the likelihood of incorrectly partitioning, enhancing the accuracy and effectiveness of the drift detection. 
\end{remark}

Subsequently, we assign $G^m$ to one of the OCCs $F$ trained for $D^b$ to characterize whether there is concept drift in $D^m$. For the samples in cluster $G_i^m$, we classify them by:
\begin{equation}
    \label{eq_judge_outlier}
    {f_i}({{\mathbf{x}}_h}) = \left\{ {\begin{array}{lll}
1,&{{\text{ if }}{{\mathbf{x}}_h} {\text{ is out-of-distribution }}}\\
0,&{{\text{ otherwise}}}
\end{array},} \right.
\end{equation}
where ${f_i}({{\mathbf{x}}_h})=0 $ indicates sample $\mathbf{x}_h$ is predicted as a normal in-cluster sample, while ${f_i}({{\mathbf{x}}_h})= 1 $ means sample $\mathbf{x}_h$ does not belong to $G_i^m$ and is considered an out-of-distribution sample.

To prevent false positives caused by outliers, we set a drift threshold $\gamma$ such that concept drift is only identified when the proportion of out-of-distribution samples is sufficiently large. $\gamma$ limits sensitivity to anomalous situations, ensuring that alerts are triggered only when a significant change occurs, thereby enhancing the robustness of the detection. Specifically, the proportion of out-of-distribution samples in $G_i^m$ is calculated as follows:
\begin{equation}
    \label{eq_compute_o}
    {{\theta }_i} = \frac{1}{{\left| {{G_i^m}} \right|}}    \sum\nolimits_{h = 1}^{\left| {{G_i^m}} \right|} {{f_i}({{\mathbf{x}}_h})}.
\end{equation}
When ${\theta}_i>\gamma$, it indicates a notable change in $G_i^m$, suggesting drift of $G_i^m$. Following such drift detection process, we obtain the drift cluster set $O$, which contains the sequential number of clusters with detected drifted samples. The concept drift detection process is summarized as Algorithm~\ref{alg_MOCD}.

Based on the learned cluster descriptors, our method detects whether concept drift occurs in the incoming chunk. More specifically, OCCs identify where concept drift occurs. To more specifically understand the concept drifts, we further define the drift sample set as: 
\begin{equation}
    \mathcal{D}^m_i = G_i^m \setminus (G_i^m \cap G_i^b), 
\end{equation}
where $\mathcal{D}^m =\{\mathcal{D}^m_i|~\mathcal{D}^m_i~\text{represents drifted samples within}~G_i^m\}$ and $G_i^m$ is one of the drifted clusters in $O$. A sample $\mathbf{x}_h\in  \mathcal{D}^m_i$ denotes that $F_i(\mathbf{x}_h) = 1$, indicating that the sample $\mathbf{x}_h$ lies within the drifted concept region.

\begin{remark}[Concept drift understanding]
    For each drift sample $\mathbf{x}_h \in  \mathcal{D}^m_i$, we compute its position relative to the nearest prototype $\mathbf{s}_{i,h}$ as:
    \begin{equation}
    \label{eq_region_vector}
    {\mathfrak{g}_h} = {\mathbf{x}_h-\mathbf{s}_{i,h}}.
    \end{equation}
    Analyzing vector $\mathbf{v}_h$ enables the assessment of the region (i.e., relative location of the drifted samples) and extent (i.e., $\|\mathfrak{g}_h\|$) of drift within each optimal cluster $G_i^m$. This analysis enables the determination of whether drift samples are concentrated in specific directions around specific prototypes, thereby outlining the shape and orientation of the drifts. 
\end{remark}

\subsection{Overall Algorithm and Complexity Analysis}

\begin{algorithm}[!t]
\SetAlgoLined 
\selectfont
\SetKwInOut{Input}{Input}\SetKwInOut{Output}{Output}

\caption{\selectfont{Cluster Descriptor-based Drift Detection}}\label{alg_MOCD}

\Input{Trained OCCs $F$, fusion queue ${\mathcal Q}^{b}$, Prototypes $S$, drift threshold $\gamma$, incoming chunk $D^m$.}
\Output{Drift cluster set $O$, drift sample set $\mathcal{D}^m$.}

Assign $D^m$ samples to $C^m$ using nearest $S$ prototypes;\\

\For{$j \leftarrow 1$ \KwTo $\kappa$}{
    \For{$i \leftarrow 1$ \KwTo $k^*$}{
        Compute $u_{i,j}$ according to Eq.~(\ref{eq_merage_subclusters});\\
        \If{$u_{i,j} = 1$}{
            $G_i^m \leftarrow G_i^m \cup C_j^m$;\\
        }
    }
}

\For{$i \leftarrow 1$ \KwTo $k^*$}{
    Calculate $\theta_i$ by Eq.~(\ref{eq_compute_o}) based on $G_i^m$ and $f_i$;\\
    \If{$\theta_i > \gamma$}{
        $O \leftarrow O \cup G_i^m$;\\
        $\mathcal{D}_i^m \leftarrow G_i^m \setminus (G_i^m \cap G_i^b)$;\\
        $\mathcal{D}^m \leftarrow \mathcal{D}^m \cup \mathcal{D}_i^m$;
    }
}

\Return{$O, \mathcal{D}^m$};
\end{algorithm}

The overall process of the proposed Imbalanced Cluster Descriptor-based Drift Detection (ICD3) can be summarized in the following steps: 1) Employ Algorithm~\ref{alg_DCDL} to perform DCDL on the base chunk $D^b$; 2) Train OCCs to describe the concepts; 3) Implement MICD according to Algorithm~\ref{alg_MOCD} to conduct interpretable drift detection on incoming chunks $D^m$ as discussed in Section~\ref{section3_MICD_drift_detection}. As a result, ICD3 detects four types of concept drift: sudden (significant changes), gradual or incremental (slow shifts over time), and recurring (reappearing concepts with different distributions). The time complexity of ICD3 is also provided below.

\begin{theorem} \label{theorem:complex}Time complexity of ICD3 is $O(\mathcal{I}*nk + n^2/k+ k\sigma)$
   \begin{proof}
    The time complexities of Imbalanced Data Learning, One-Cluster Classifier Learning, and Drift Detection are $O(nlogn+\mathcal{I}*nk + k^2logk + n^2/k)$, $O(k\sigma)$, and $O(n + k)$, respectively. $\mathcal{I}$ and $k$ represents the maximum number of iterations and the number of sub-clusters, respectively. $n=\max(n^b,n^m)$ ensures that the time complexity is analyzed in the worst case. $\sigma$ denotes the complexity of a specific OCC. Accordingly, the overall time complexity of ICD3 is $O(\mathcal{I}*nk + k^2logk + n^2/k+ nlogn+ k\sigma)$. As $k$ is usually a very small value compared to $n$ satisfying $k\ll n$, $n^2/k$ dominates both $nlogn$ and $k^2logk$. Accordingly, the overall time complexity can be simplified to $O(\mathcal{I}*nk + n^2/k+ k\sigma)$.
\end{proof}
\end{theorem}

\section{Experiment}

\subsection{Experimental Settings}\label{section4: experiment setup}

\subsubsection{Experimental Design}

Four types of experiments have been conducted:
\begin{itemize}
    \item \textit{Comparative Study of Drift Detection Performance}: To demonstrate the superiority of ICD3, we compare its drift detection accuracy, area under the ROC curve, and G-Mean against conventional and state-of-the-art techniques across various datasets with different imbalance ratios. 
    \item \textit{Ablation Studies}: To specifically demonstrate the effectiveness of ICD3 design, we ablate its different modules to form corresponding versions for comparison.
    \item \textit{Drift Understanding Capability Evaluation}: Qualitative visual experiments have been conducted to intuitively show the capability of ICD3 for precisely identifying drifted clusters, as well as its effectiveness in reflecting the distribution pattern of the drifted samples. 
    \item \textit{Parameter Sensitivity Study}: To illustrate the generality and practicality of the ICD3 across different scenarios, we conduct sensitivity testing on the drift threshold parameter employed in the ICD3. 
\end{itemize}

\subsubsection{Evaluation Metrics}
Accuracy~\cite{acc}, G-Mean~\cite{lid}, and AUC~\cite{auc}, are adopted to facilitate multi-aspect assessment of drift detection performance, taking into account both the overall accuracy and the impact of data imbalance.

\begin{table}[!t]
    \caption{Dataset statistics. $n$, $k$, and $k'$ represent the number of samples, clusters, and sampled cluster counts, respectively. $\cal M$ is the count ratio of large clusters to small clusters.}
    \label{tb_real_configure}
    \centering
    \begin{tabular}{l|c|ccccc}
    \toprule
     \multicolumn{2}{c|}{Dataset}&Feature   &$n$    &$k$    &$k'$  &$\cal M$  \\ 
     \midrule
     \multirow{7}{*}{Real}&Avila      &10 &20867  &12 &3  &2:1    \\
     &Covtype    &10 &581012 &4  &3  &2:1    \\
     &Insect     &49 &5325   &5  &5  &3:2    \\
     &Noaa       &9  &18159  &2  &2  &1:1    \\
     &Occupy     &4  &20560  &2  &3  &2:1    \\
     &Posture    &3  &164860 &11 &4  &2:2    \\
     &Shuttle    &9  &43500  &7  &2  &1:1    \\
     \midrule
     \multirow{7}{*}{Synthetic}&2D-2G-M&2&250000&2&2&1:1\\
     &2D-2G-C& 2& 250000& 2& 2& 1:1\\
     &2D-2G-V& 2& 250000& 2& 2& 1:1\\
     &2D-4G-M& 2& 250000& 4& 4& 2:2\\
     &2D-4G-C& 2& 250000& 4& 4& 2:2\\
     &2D-8G-M& 2& 1000000& 8& 8& 6:2\\
     &2D-M-N& 2& 250000& 2&  2& 1:1\\
     \bottomrule
    \end{tabular}
\end{table}

\subsubsection{Datasets}

Seven real benchmark datasets~\cite{uci} and seven synthetic datasets are employed for the experiments. Their dataset statistics are demonstrated in Table~\ref{tb_real_configure}. Each dataset has been prepared with a relatively large base chunk for initial distribution learning and has been divided into 500 data chunks, including 250 with the same distribution as the base chunk, and the remainder 250 with drifted imbalanced concepts generated. Given that our work is the first attempt to address the problem of imbalanced concept drift detection under the unsupervised scenario, we design such a generator algorithm to simulate relatively extreme cases of the imbalanced concepts and drifts to more comprehensively evaluate the proposed method. We set the imbalance ratio at $\text{IR}=|C_l|/|C_s|=15$, indicating that the size of the largest cluster is 15 times larger than the smallest one. Then we discuss how to generate drifted streaming data chunks with imbalanced concepts for real benchmark and synthetic datasets, respectively. Each dataset, except for the 2D-8G-M dataset, has a base chunk of 2000 samples, while the incoming chunk contains 500 samples. For the 2D-8G-M dataset, the base chunk consists of 4000 samples, and the incoming chunk contains 2000 samples.

\begin{table*}[!t]

\centering
\scriptsize
\caption{Accuracy, AUC, and G-Mean results of drift detection. \textbf{Bold} indicates the optimal result, and \underline{underline} indicates the second-best result. Using accuracy as an example, the row of ``Ave. Acc" represents the average accuracy of each method, and the ``Ave. Rank'' row reports the average accuracy rankings. The same interpretation holds for AUC and G-Mean.}
\label{tb_drift_detection_acc}

\resizebox{2.06\columnwidth}{!}{\begin{tabular}{l|l|c|cccccc|cc}
\toprule
Metric&\multicolumn{2}{c|}{Dataset}&QT-EWMA &EI-KMeans &OCDD &Qtree & MWW & MCD&OICD3 (ours)  &MICD3 (ours) \\ 
\midrule
\multirow{16}{*}{ACC} &\multirow{7}{*}{Synthetic} & 2D-2G-M &0.5600$\pm$0.0493 &0.5061$\pm$0.0000 &0.5000$\pm$0.0000 &0.5391$\pm$0.0271 &0.4862$\pm$0.0000 &0.5002$\pm$0.0006&\textbf{0.7663$\pm$0.0000} &\underline{0.7260$\pm$0.0000} \\
&&2D-2G-C &0.5120$\pm$0.0223 &0.5040$\pm$0.0000 &0.5000$\pm$0.0000 &0.4911$\pm$0.0174 &0.5060$\pm$0.0000 &0.5000$\pm$0.0000&\textbf{0.9322$\pm$0.0000} &\underline{0.9243$\pm$0.0000} \\
&&2D-2G-V &0.5180$\pm$0.0351 &0.5540$\pm$0.0000 &0.5000$\pm$0.0000 &0.4961$\pm$0.0200 &0.4782$\pm$0.0000 &0.5025$\pm$0.0078 &\underline{0.9100$\pm$0.0000} &\textbf{0.9282$\pm$0.0000} \\
&&2D-4G-M &0.5043$\pm$0.0145 &0.5042$\pm$0.0000 &0.5000$\pm$0.0000 &0.4933$\pm$0.0282 &0.4981$\pm$0.0000 &0.5000$\pm$0.0000&\textbf{0.6100$\pm$0.0081} &\underline{0.5690$\pm$0.0083} \\
&&2D-4G-C &0.4970$\pm$0.0236 &0.5220$\pm$0.0000 &0.5000$\pm$0.0000 &0.5140$\pm$0.0151 &0.4980$\pm$0.0000 &0.5000$\pm$0.0000&\textbf{0.7960$\pm$0.0172} &\underline{0.6580$\pm$0.0304} \\
&&2D-8G-M &0.5164$\pm$0.0187 &0.5000$\pm$0.0000 &0.5000$\pm$0.0000 &0.5050$\pm$0.0154 &0.5000$\pm$0.0000 &0.5000$\pm$0.0000 &\textbf{0.7510$\pm$0.0453} &\underline{0.7141$\pm$0.0273} \\
&&2D-M-N &0.5000$\pm$0.0155 &0.5000$\pm$0.0000 &0.5000$\pm$0.0000 &0.5011$\pm$0.0144 &0.5000$\pm$0.0000 &0.5000$\pm$0.0000 &\textbf{0.9182$\pm$0.0161} &\underline{0.7753$\pm$0.0211} \\
\cmidrule(lr){2-11}
&\multirow{7}{*}{Real} &Avila &0.5040$\pm$0.0103 &0.5100$\pm$0.0000 &0.5000$\pm$0.0000 &0.5000$\pm$0.0092 &0.5000$\pm$0.0000 &0.5000$\pm$0.0000&\textbf{0.5500$\pm$0.0194} &\underline{0.5250$\pm$0.0071} \\
&&Covtype &\underline{0.5141$\pm$0.0193} &0.5041$\pm$0.0000 &0.5000$\pm$0.0000 &0.5112$\pm$0.0192 &0.5022$\pm$0.0000 &0.5000$\pm$0.0000 &\textbf{0.5174$\pm$0.0085} &0.4983$\pm$0.0104 \\
&&Insect &0.5060$\pm$0.0133 &\underline{0.5240$\pm$0.0000} &0.5000$\pm$0.0000 &0.5073$\pm$0.0135 &0.5000$\pm$0.0000 &0.5000$\pm$0.0000&\textbf{0.5340$\pm$0.0113} &0.5142$\pm$0.0123 \\
&&Noaa &0.5993$\pm$0.0527 &\underline{0.9060$\pm$0.0000} &0.5000$\pm$0.0000 &0.5390$\pm$0.0271 &0.5100$\pm$0.0000 &0.5002$\pm$0.0006&0.8750$\pm$0.0946 &\textbf{0.9981$\pm$0.0044} \\
&&Occupy &0.5000$\pm$0.0000 &0.6820$\pm$0.0000 &0.5000$\pm$0.0000 &0.5000$\pm$0.0000 &0.5063$\pm$0.0000 &0.5000$\pm$0.0000&\underline{0.8052$\pm$0.0544} &\textbf{0.9433$\pm$0.0223} \\
&&Posture &\underline{0.5070$\pm$0.0155} &0.4920$\pm$0.0000 &0.5000$\pm$0.0000 &0.5000$\pm$0.0178 &0.5000$\pm$0.0000 &0.5002$\pm$0.0006 &\textbf{0.5112$\pm$0.0170} &0.5053$\pm$0.0084 \\
&&Shuttle &0.5000$\pm$0.0000 &0.5000$\pm$0.0000 &0.5000$\pm$0.0000 &0.5000$\pm$0.0011 &0.5000$\pm$0.0000 &0.5002$\pm$0.0006&\underline{0.7852$\pm$0.0350} &\textbf{0.9944$\pm$0.0020} \\
\midrule
\multicolumn{3}{c|}{Ave. Acc} &0.5170 &0.5506 &0.5000 &0.5069 &0.4989 &0.5002&\underline{0.7330} &\textbf{0.7338} \\
\multicolumn{3}{c|}{Ave. Rank} &4.250 &4.321 &6.393 &5.357 &6.214 &5.821&\textbf{1.357} &\underline{2.286}\\ 

\midrule

\multirow{16}{*}{AUC} &\multirow{7}{*}{Synthetic} &2D-2G-M& 0.5223$\pm$0.0561	& 0.4730$\pm$0.0000	& 0.7066$\pm$0.0000	& 0.5112$\pm$0.0191	& 0.4984$\pm$0.0000	& 0.6029$\pm$0.0118	& \textbf{0.8312$\pm$0.0000}	& \underline{0.8307$\pm$0.0000}\\	
&&2D-2G-C& 0.5039$\pm$0.0176	& 0.4702$\pm$0.0000	& 0.8236$\pm$0.0000	& 0.5170$\pm$0.0126	& 0.5116$\pm$0.0000	& 0.5359$\pm$0.0314	& \textbf{0.9821$\pm$0.0000}	& \underline{0.9668$\pm$0.0000}\\	
&&2D-2G-V& 0.5130$\pm$0.0291	& 0.7438$\pm$0.0000	& 0.8457$\pm$0.0000	& 0.5170$\pm$0.0109	& 0.4375$\pm$0.0000	& 0.6317$\pm$0.0178	& \textbf{0.9825$\pm$0.0000}	& \underline{0.9717$\pm$0.0000}\\	
&&2D-4G-M& 0.5037$\pm$0.0354	& 0.5613$\pm$0.0000	& 0.4499$\pm$0.0000	& 0.5328$\pm$0.0164	& 0.4437$\pm$0.0000	& 0.4825$\pm$0.0148	& \underline{0.7309$\pm$0.0168}	& \textbf{0.8012$\pm$0.0080}\\	
&&2D-4G-C& 0.4837$\pm$0.0193	& 0.5479$\pm$0.0000	& 0.7266$\pm$0.0000	& 0.5032$\pm$0.0042	& 0.4893$\pm$0.0000	& 0.4915$\pm$0.0183	& \textbf{0.9471$\pm$0.0046}	& \underline{0.9335$\pm$0.0044}\\	
&&2D-8G-M& 0.5030$\pm$0.0179	& 0.5009$\pm$0.0000	& 0.5780$\pm$0.0000	& 0.4918$\pm$0.0061	& 0.5306$\pm$0.0000	& 0.5180$\pm$0.0284	& \textbf{0.8183$\pm$0.0313}	& \underline{0.7805$\pm$0.0403}\\	
&&2D-M-N& 0.5187$\pm$0.0186	& 0.6503$\pm$0.0000	& 0.4403$\pm$0.0000	& 0.4988$\pm$0.0040	& 0.5548$\pm$0.0000	& 0.5375$\pm$0.0317	& \textbf{0.9653$\pm$0.0015}	& \underline{0.8097$\pm$0.0064}\\	
\cmidrule(lr){2-11}
&\multirow{7}{*}{Real} &Avila& 0.5150$\pm$0.0234	& 0.4994$\pm$0.0000	& \textbf{0.7689$\pm$0.0000}	& 0.5044$\pm$0.0099	& 0.5504$\pm$0.0000	& 0.5179$\pm$0.0245	& \underline{0.5667$\pm$0.0193}	& 0.5236$\pm$0.0119\\	
&&Covtype& 0.5072$\pm$0.0302	& 0.4597$\pm$0.0000	& 0.3304$\pm$0.0000	& \underline{0.5100$\pm$0.0073}	& \textbf{0.5539$\pm$0.0000}	& 0.4754$\pm$0.0341	& 0.5045$\pm$0.0286	& 0.4703$\pm$0.0251\\	
&&Insect& 0.5299$\pm$0.0315	& \textbf{0.7661$\pm$0.0000}	& 0.6236$\pm$0.0000	& 0.5070$\pm$0.0060	& 0.4672$\pm$0.0289	& 0.5322$\pm$0.0946	& 0.5663$\pm$0.0243	& \underline{0.6754$\pm$0.0242}\\	
&&Noaa& 0.5837$\pm$0.0442	& 0.9683$\pm$0.0000	& \underline{0.9993$\pm$0.0000}	& 0.7010$\pm$0.1386	& 0.4982$\pm$0.0000	& 0.7831$\pm$0.0185	& 0.9857$\pm$0.0119	& \textbf{1.0000$\pm$0.0000}\\	
&&Occupy& 0.5345$\pm$0.0255	& 0.7738$\pm$0.0000	& 0.7988$\pm$0.0000	& 0.5606$\pm$0.0282	& 0.5009$\pm$0.0000	& 0.6036$\pm$0.0272	& \underline{0.9835$\pm$0.0273}	& \textbf{0.9948$\pm$0.0029}\\	
&&Posture& 0.5036$\pm$0.0198	& 0.5012$\pm$0.0000	& 0.4939$\pm$0.0000	& 0.4944$\pm$0.0057	& \underline{0.5159$\pm$0.0000}	& 0.4973$\pm$0.0247	& 0.5117$\pm$0.0234	& \textbf{0.5320$\pm$0.0105}\\	
&&Shuttle& 0.5024$\pm$0.0226	& 0.4812$\pm$0.0000	& 0.7213$\pm$0.0000	& 0.4998$\pm$0.0074	& 0.5012$\pm$0.0000	& 0.7647$\pm$0.0489	& \underline{0.9647$\pm$0.0059}	& \textbf{0.9999$\pm$0.0000}\\	
\midrule
\multicolumn{3}{c|}{Ave. AUC} &0.5160 &0.5998 &0.6648 &0.5249 &0.5038 &0.5696&\textbf{0.8100} & \underline{0.8064} \\
\multicolumn{3}{c|}{Ave. Rank} &5.857&	5.286&	4.214&5.929&	5.700&	4.929&	\textbf{2.000}&	\textbf{2.071}\\
\midrule

\multirow{16}{*}{G-Mean} &\multirow{7}{*}{Synthetic} &2D-2G-M& 0.5151$\pm$0.0496	& 0.1972$\pm$0.0000	& 0.0000$\pm$0.0000	& 0.3325$\pm$0.0559	& 0.0620$\pm$0.0000	& 0.0063$\pm$0.0190	 &\textbf{0.7460$\pm$0.0000}	& \underline{0.7182$\pm$0.0000}\\	
&&2D-2G-C& 0.4910$\pm$0.0183	& 0.2385$\pm$0.0000	& 0.0000$\pm$0.0000	& 0.2994$\pm$0.0457	& 0.1540$\pm$0.0000	& 0.0000$\pm$0.0000	& \textbf{0.9316$\pm$0.0000}	& \underline{0.9230$\pm$0.0000}\\	
&&2D-2G-V& 0.5035$\pm$0.0236	& 0.3443$\pm$0.0000	& 0.0000$\pm$0.0000	& 0.2553$\pm$0.0583	& 0.1242$\pm$0.0000	& 0.0228$\pm$0.0684	& \underline{0.9086$\pm$0.0000}	& \textbf{0.9280$\pm$0.0000}\\	
&&2D-4G-M& \underline{0.5087$\pm$0.0246}	& 0.1537$\pm$0.0000	& 0.0000$\pm$0.0000	& 0.2936$\pm$0.0594	& 0.0000$\pm$0.0000	& 0.0000$\pm$0.0000	& \textbf{0.5639$\pm$0.0183}	& 0.3911$\pm$0.0232\\	
&&2D-4G-C& 0.4859$\pm$0.0208	& 0.2645$\pm$0.0000	& 0.0000$\pm$0.0000	& 0.1597$\pm$0.0262	& 0.0000$\pm$0.0000	& 0.0000$\pm$0.0000	& \textbf{0.7788$\pm$0.0228}	& \underline{0.5693$\pm$0.0556}\\	
&&2D-8G-M& 0.2619$\pm$0.0193	& 0.0000$\pm$0.0000	& 0.0000$\pm$0.0000	& 0.2001$\pm$0.0216	& 0.0000$\pm$0.0000	& 0.0000$\pm$0.0000	& \textbf{0.7283$\pm$0.0629}	& \underline{0.6989$\pm$0.0471}\\	
&&2D-M-N& 0.5045$\pm$0.0202	& 0.0000$\pm$0.0000	& 0.0000$\pm$0.0000	& 0.0514$\pm$0.0366	& 0.0000$\pm$0.0000	& 0.0000$\pm$0.0000	& \textbf{0.9177$\pm$0.0157}	& \underline{0.7491$\pm$0.0328}\\	
\cmidrule(lr){2-11}
&\multirow{7}{*}{Real}&Avila& 0.0537$\pm$0.1612	& 0.1882$\pm$0.0000	& 0.0000$\pm$0.0000	& 0.2021$\pm$0.0492	& 0.0000$\pm$0.0000	& 0.0000$\pm$0.0000	& \textbf{0.4728$\pm$0.1155}	& \underline{0.3451$\pm$0.0136}\\	
&&Covtype& \underline{0.4293$\pm$0.0576}	& 0.2147$\pm$0.0000	& 0.0000$\pm$0.0000	& 0.2237$\pm$0.0533	& 0.0632$\pm$0.0000	& 0.0000$\pm$0.0000	& \textbf{0.4376$\pm$0.0925}	& 0.1962$\pm$0.0483\\	
&&Insect& \textbf{0.5194$\pm$0.0263}	& 0.2191$\pm$0.0000	& 0.0000$\pm$0.0000	& 0.1220$\pm$0.0433	& 0.0000$\pm$0.0000	& 0.0000$\pm$0.0000	& \underline{0.4776$\pm$0.0511}	& 0.2042$\pm$0.1125\\	
&&Noaa& 0.5180$\pm$0.0770	& \underline{0.9027$\pm$0.0000}	& 0.0000$\pm$0.0000	& 0.6064$\pm$0.2101	& 0.1756$\pm$0.0000	& 0.0089$\pm$0.0268	& 0.8595$\pm$0.1122	& \textbf{0.9984$\pm$0.0039}\\	
&&Occupy& 0.0000$\pm$0.0000	& 0.6514$\pm$0.0000	& 0.0000$\pm$0.0000	& 0.3845$\pm$0.0680	& 0.1411$\pm$0.0000	& 0.0000$\pm$0.0000	& \underline{0.7802$\pm$0.0637}	& \textbf{0.9413$\pm$0.0234}\\	
&&Posture& \textbf{0.5009$\pm$0.0131}	& 0.1245$\pm$0.0000	& 0.0000$\pm$0.0000	& 0.1208$\pm$0.0622	& 0.0631$\pm$0.0000	& 0.0089$\pm$0.0268	& \underline{0.2550$\pm$0.1798}	& 0.1305$\pm$0.1322\\	
&&Shuttle& 0.0000$\pm$0.0000	& 0.1400$\pm$0.0000	& 0.0000$\pm$0.0000	& 0.1290$\pm$0.0527	& 0.0000$\pm$0.0000	& 0.0063$\pm$0.0190	& \underline{0.7541$\pm$0.0473}	& \textbf{0.9940$\pm$0.0016}\\	
\midrule
\multicolumn{3}{c|}{Ave. G-Mean} &0.3780&	0.2599&	0.0000	&0.2415&	0.0560&	0.0038&	\textbf{0.6866}&\underline{0.6277} \\
\multicolumn{3}{c|}{Ave. Rank} &3.077&	4.077&	6.615&	4.154&	5.769	&6.308&	\textbf{1.462}&	\underline{2.308}\\

\bottomrule

\end{tabular}}

\vspace{1mm}

\end{table*}

For the synthetic datasets, the imbalanced and drifted data streams are simulated through a two-step procedure: 1) Generate base chunks with multiple clusters of varying sample sizes according to a specified imbalance ratio (i.e., IR = 15); 2) Induce concept drift by modifying the distribution of minority clusters, thereby simulating the challenging imbalanced concept drifting scenarios. The synthetic datasets include both Gaussian and non-Gaussian distributions.
For Gaussian datasets, drift is introduced by altering the mean vector, covariance matrix, or both of selected clusters. A random drift margin $u$ is selected from the interval $(0.1, 1)$ to control the extent of distributional change. For example, a mean drift shifts the cluster centroid, while a covariance drift changes its shape.
For non-Gaussian datasets, we simulate drift by adding noise samples around the boundaries of moon-shaped clusters, mimicking structure-preserving but noisy concept shifts. These datasets are particularly suitable for testing the model's capability in detecting drift in complex distributions.
For real datasets, samples within different benchmark classes are naturally imbalanced.
After obtaining imbalanced data chunks through the IR-guided random sampling, we introduce drifts by changing the order of the sample dimensions for a part of the samples and introducing samples from other classes following the configurations in~\cite{cbld}.

\subsubsection{Counterparts}

The proposed ICD3 is compared with six counterparts, including QT-EWMA~\cite{QT-EWKM}, EI-Kmeans~\cite{EI-Kmeans}, OCDD~\cite{OCDD}, QTree~\cite{QuanTree}, Mann-Whitney-Wilcoxon (MWW) test~\cite{MWW}, and MCD~\cite{mcd}. The parameters for these methods are set based on the original papers and released code. To ensure the reliability of the experiments and minimize the introduction of additional factors, We combine with Classical OCSVM~\cite{OCSVM} as OICD3, and combine with the latest ME\_SVDD~\cite{ME_SVDD} as MICD3, providing two different versions of our method.  
In all experiments, we set the drift threshold $\gamma = 0.2$ for consistency.

\begin{figure}[!t]
\centering
\includegraphics[width=3.5in]{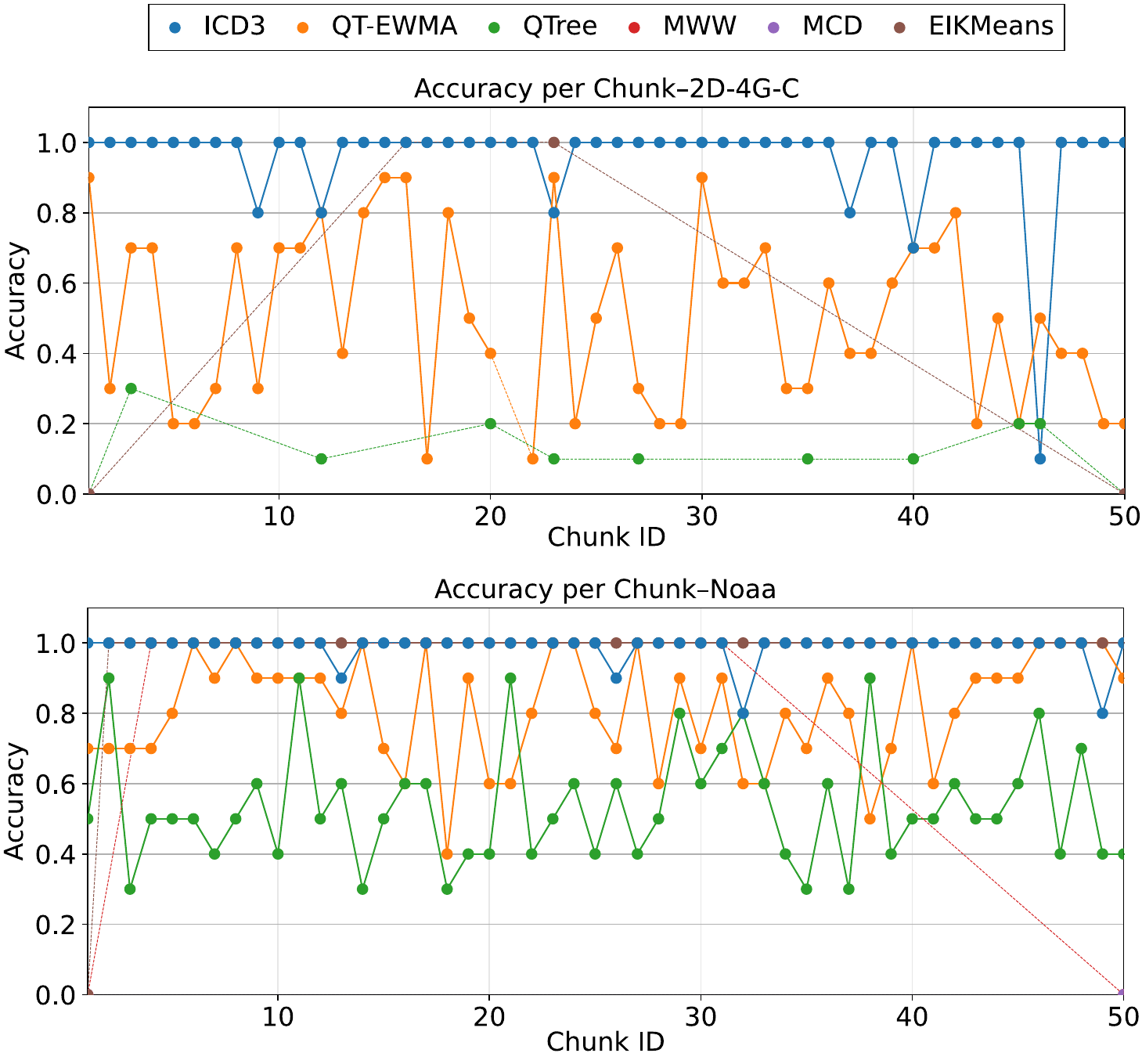}
\caption{Accuracy across chunks on 2D-4G-C and Noaa dataset.}
\label{fig_chunkacc}
\end{figure}

\subsection{Comparative Study of Drift Detection Performance}
\label{section4: accuracy_tab}
This section investigates the drift detection performance of different algorithms and statistically analyze the superiority of OICD3 and MICD3.
\subsubsection{Overall Drift Detection Performance}
Drift Detection Performance of different methods are compared in 
Table~\ref{tb_drift_detection_acc} w.r.t accuracy. The best and second-best results on each dataset are highlighted in \textbf{bold} and \underline{underline}, respectively. The observations include the following five aspects: 1) Overall, both versions of our method, OICD3, and MICD3, perform best or second-best on almost all datasets, indicating its superiority in drift detection. 2) Although OICD3 does not have the best accuracy performance on the Noaa dataset, it maintains the second-best and is only marginally outperformed by the leading models. 3) OICD3 and MICD3 differ only in their selection of descriptors, yet both effectively detect drift in imbalanced data, demonstrating the robustness of the method selection and exhibiting high accuracy in drift detection. 4) It can be seen that the drift detection accuracy of OCDD is always 0.5 on different datasets. This is because that OCDD is primarily designed for global concept drift detection, and thus completely fails in detecting drifts of certain clusters in an imbalanced data chunk. Since it always classifies data samples as normal ones here, its detection accuracy is always 0.5. 5) It can be observed that the OICD3 outperforms the ME\_SVDD-based MICD3 on more datasets. Consequently, we adopt the OICD3 version for all subsequent experiments, and for simplicity, we refer to it directly as ICD3.

AUC and G-Mean Performance of different methods are compared in Table~\ref{tb_drift_detection_acc}. The observations include the following two aspects: 1) Overall, both versions of our ICD3 method, i.e., OICD3 and MICD3, perform best or second-best on almost all datasets in terms of both AUC and G-Mean metrics. This indicates their superiority in drift detection. 2) Although OCDD demonstrates relatively high AUC scores on certain datasets, its practical drift detection appears to be constrained in specific scenarios. Specifically, OCDD is primarily designed for global concept drift detection. Consequently, when drift occurs in only a small subset of the samples, OCDD tends to regard these subtle shifts as part of the normal variation, making it challenging to detect in this scenario. Significance study~\cite{r1test} is conducted based on ACC performance in Table~\ref{tb_drift_detection_acc}. OICD3 shows statistically significant superiority over all other methods except MICD3.

\subsubsection{Streaming Chunk-Wise Drift Detection Accuracy}
Detailed drift detection accuracy on a series of incoming data chunks is also demonstrated to enable a better understanding of how each algorithm responds to drift events over time. The result on the 2D-4G-C and Noaa datasets can be seen in Fig.~\ref{fig_chunkacc}. The dashed lines indicate that no drift was detected by the method in these chunks. OCDD completely fails in detecting drifts in an imbalanced data chunk, so we omit the results of OCDD. It can be observed that ICD3 demonstrates strong performance across nearly all chunks, indicating its robustness and timeliness.

\begin{figure}[!t]
\centering
\includegraphics[width=3.5in]{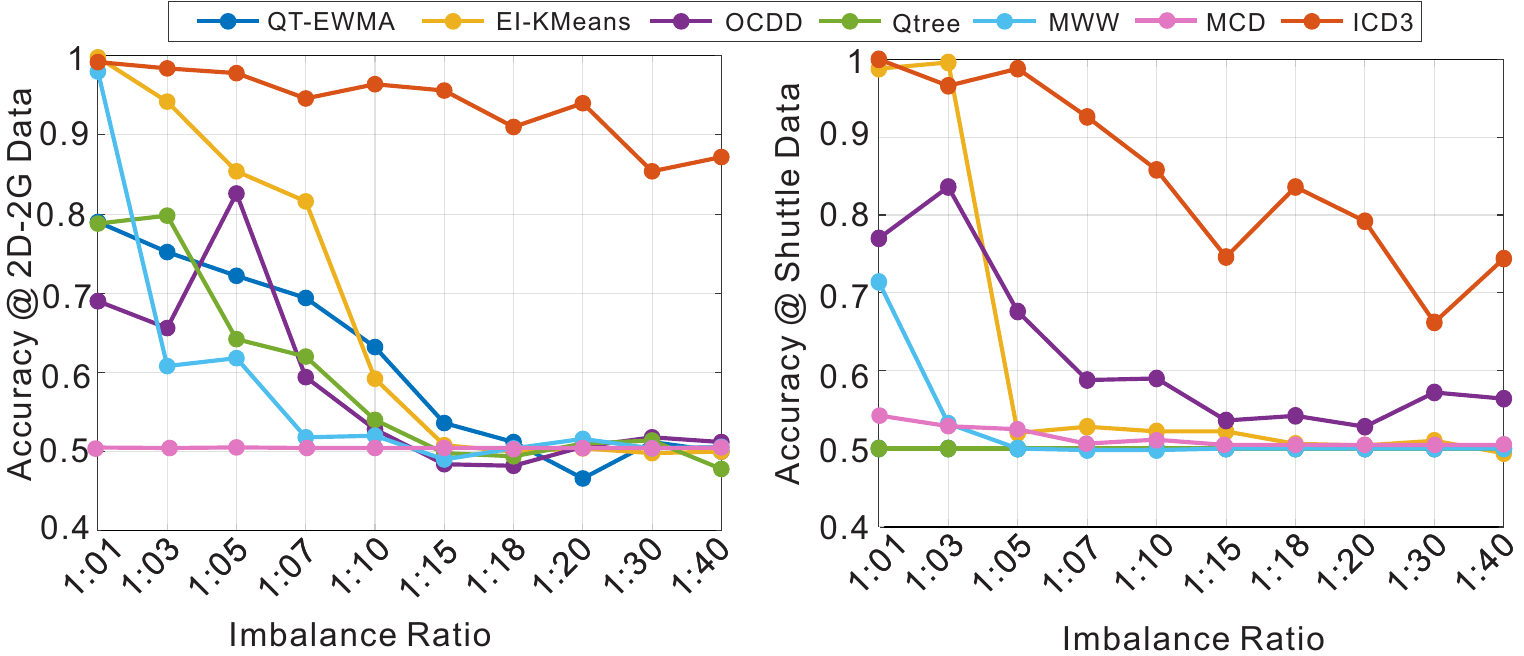}
\caption{Comparison of the Accuracy of ICD3 and counterparts across different imbalance ratios.}
\label{fig_Differ_IR}
\end{figure}

\subsubsection{Drift Detection Performance on Varying Imbalance Rate}

To further illustrate the robustness of our method across different degrees of cluster imbalance, we construct ten imbalanced datasets from the synthetic dataset 2D-2G and another ten from the real dataset Shuttle, respectively, using the same set of imbalance ratios $\mathcal{IR} =\{1,3,5,7,10,15,18,20,30,40\} $. The results are presented in Fig.~\ref{fig_Differ_IR}. As can be seen, our ICD3 is superior to its counterparts across different imbalance ratios. More specifically, on both real or synthetic datasets, the accuracy of ICD3 and the other five comparison methods decreases with the imbalance ratio increasing. However, ICD3 maintains a remarkable level of accuracy while the other methods show notable decreases. This superiority is attributed to the mechanism of ICD3 that utilizes multiple independent descriptors to effectively describe clusters of different sizes.

\subsection{Ablation Study}
\label{section4: abs}
\begin{figure}[!t]
    \centering
    \includegraphics[width=3.5in]{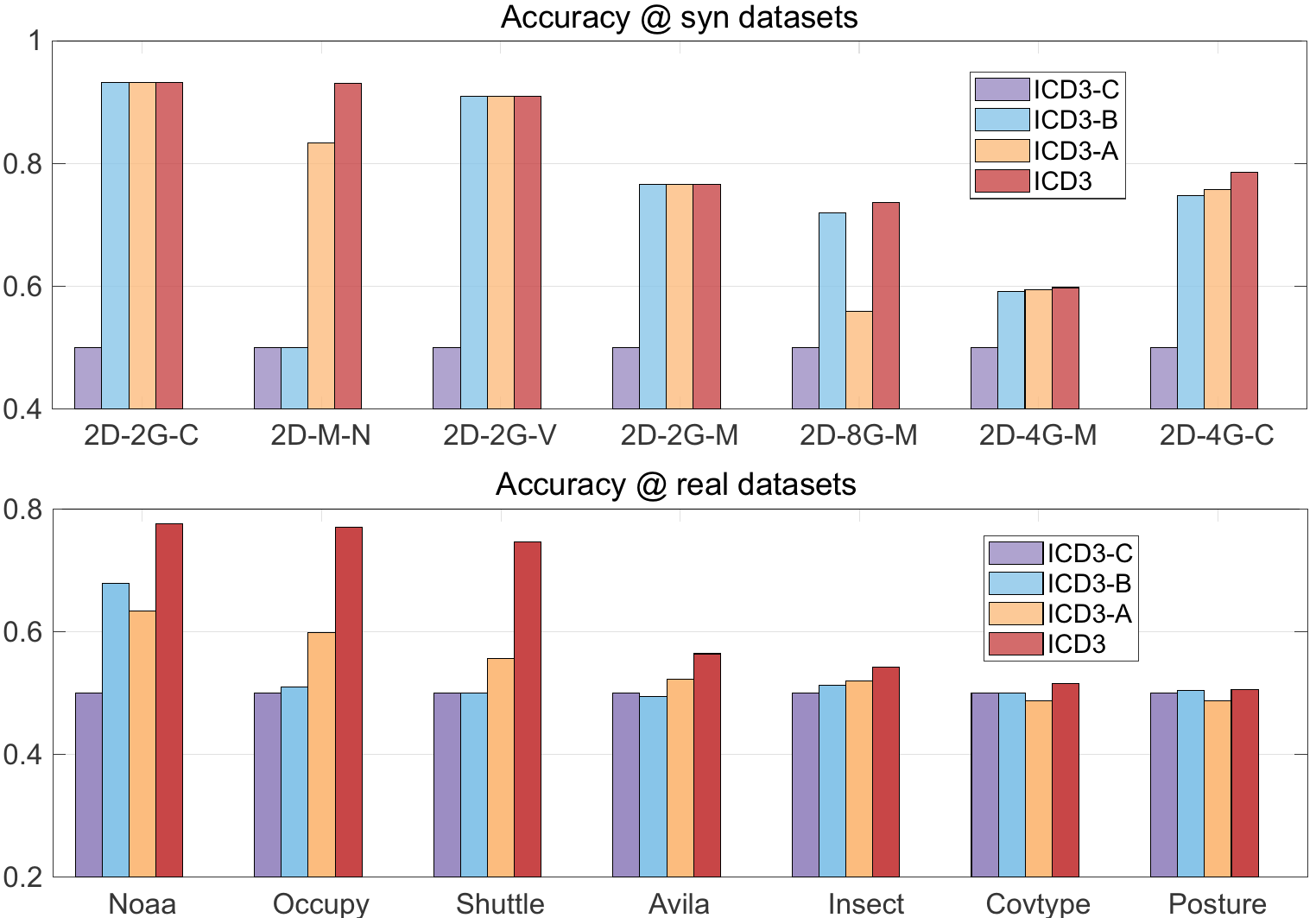}
    \caption{Drift detection performance of ICD3 and its three ablated versions (i.e., ICD3-A, ICD3-B, and ICD3-C) on all the 14 datasets. 
    }
    \label{fig:ab_study}
\end{figure}

To explicitly demonstrate the effectiveness of the core components of ICD3, three ablated versions are compared in Fig.~\ref{fig:ab_study}. Three variants are summarized as follows:
1) To evaluate the effectiveness of the density-guided initialization method, we compare ICD3 with its variant, ICD3-A, which employs the traditional strategy of randomly selecting samples as initial prototypes for clustering. 2) To evaluate the effectiveness of DCDL, we further modify the mechanism in ICD3-A to utilize traditional $k$-means, yielding ICD3-B. 3) To evaluate the effectiveness of using OCCs to track the distribution changes of each cluster individually, we use only one OCC to describe the distribution changes of all clusters, thus forming ICD3-C.

It can be observed from Fig.~\ref{fig:ab_study} that the performance of ICD3 is superior to its three ablated variants. More specific observations are three-fold: 1) ICD3 outperforms ICD3-A on 12 out of 14 datasets, this indicates the necessity to apply an enhanced center selection scheme when dealing with imbalanced data. 2) The performance of ICD3-A is not worse than ICD3-B on 10 datasets, validating the importance of the DCDL mechanism. 3) ICD3-B outperforms ICD3-C on 10 datasets, proving that multiple descriptors are more effective in identifying concept changes specific to individual clusters than using only one descriptor. To sum up, ICD3 surpasses the three ablated versions on 11 datasets and shows similar performance on the remaining 3, highlighting the necessity of all the proposed modules.

\subsection{Evaluation of Drift Understanding Capability}
\label{section4: Intuitive r}

\begin{figure}[!t]	
\centerline{\includegraphics[width=3.6in]{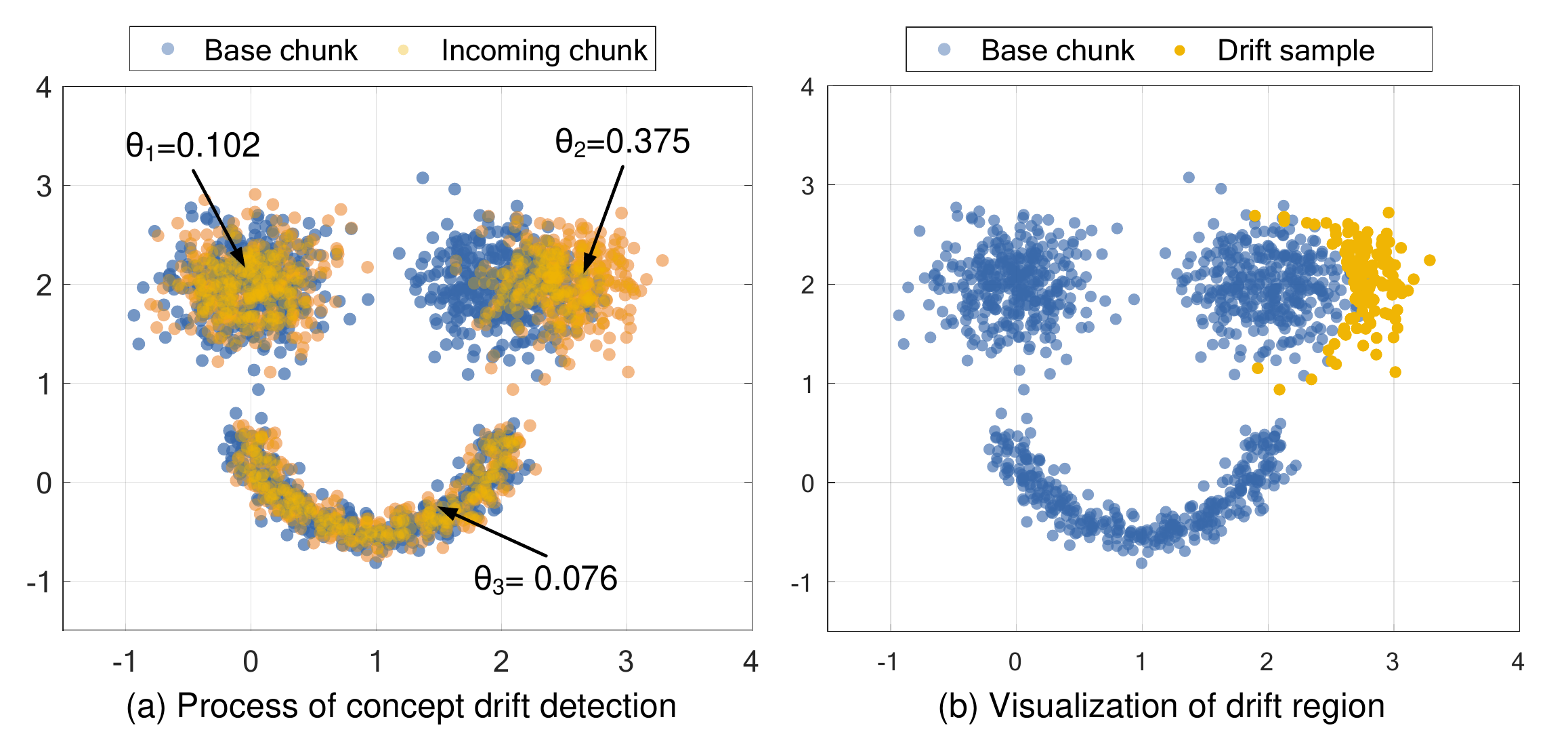}}
\caption{Interpretable drift detection process of ICD3. (a) illustrates the spatial distribution of samples in two chunks indicated by blue and yellow colors. $\theta_i$ represents the proportion of out-of-distribution samples in the $i$-th cluster of the incoming chunk, $\theta_i > \gamma $ indicates that the specific cluster occurs drift. (b) demonstrates that ICD3 accurately pinpointed the drift region and corresponding samples within the incoming chunk.}
\label{fig_intutive_study}
\end{figure}

To illustrate the drift detection capability of ICD3, we generate base chunks and incoming chunks on the smiley face dataset, inducing drift in one of the clusters in the incoming chunk and using ICD3 to detect drift. The sample distribution with drift detection processes is visualized in Fig.~\ref{fig_intutive_study} (a) and (b). Specifically, the observations can be summarized as follows: 1) Fig.~\ref{fig_intutive_study} (a) indicates that ICD3 detects a sample out-of-distribution proportion in each cluster of the incoming data chunk, where $\theta > \gamma$. This shows that ICD3 successfully identifies that concept drift occurs. 2) The sample out-of-distribution proportion for the second cluster is $\theta_2 = 0.375 $, which is exceeding the threshold $\gamma=0.2$, indicating that this cluster occurs concept drift. 3) As shown in Fig.~\ref{fig_intutive_study} (b), ICD3 can precisely locate the drift region and specifically illustrate the characteristics of this area. In summary, ICD3 can informatively reveal whether concept drift occurs, where it occurs, and what the drifted regions look like.

\begin{figure}[!t]
\centering
\includegraphics[width=3.5in,height=5in]{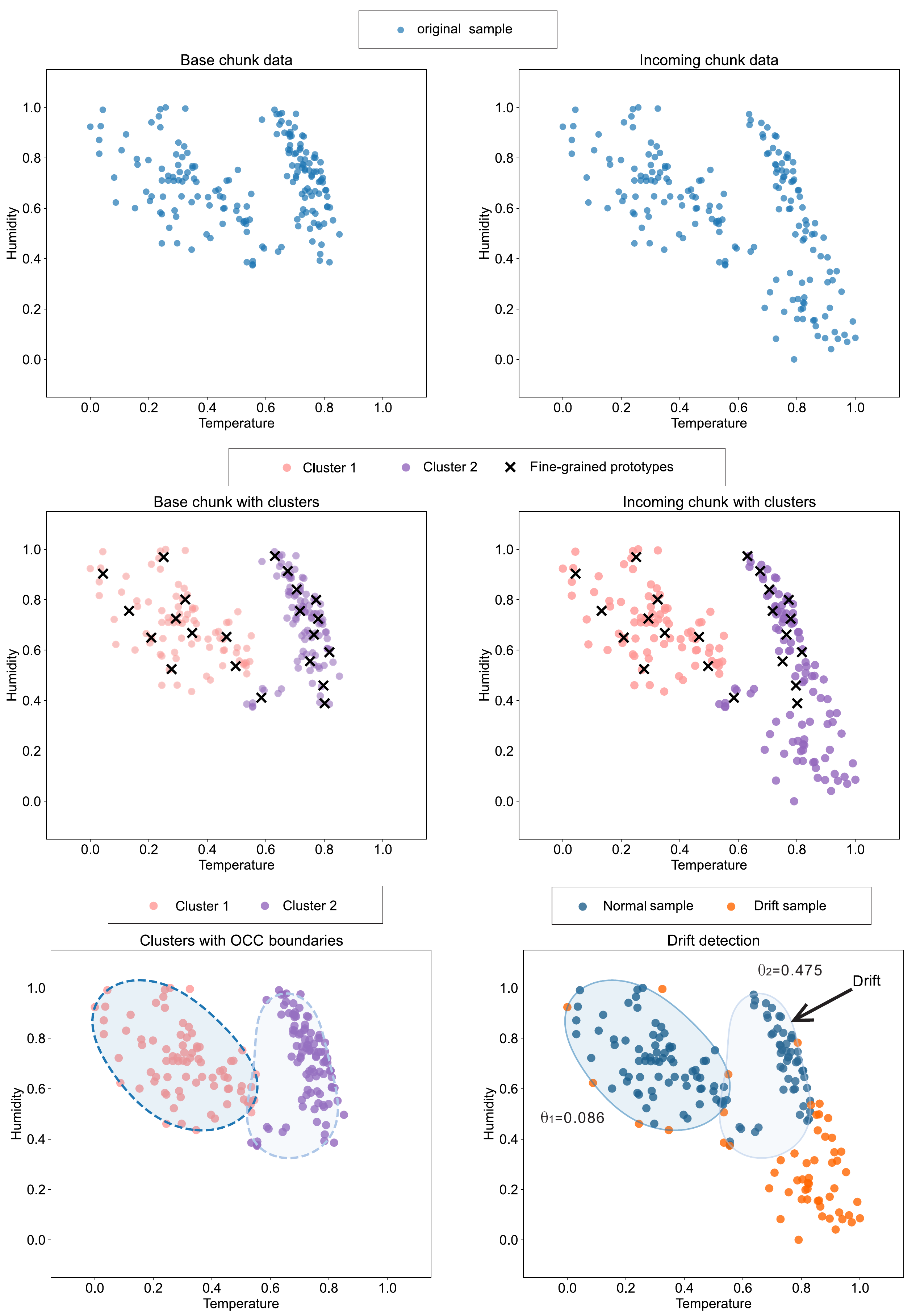}
\caption{Demonstration on the real-world Climate dataset. On the left column, the base chunk is depicted, where ICD3 learns fine-grained prototypes and captures the distribution of two clusters, and subsequently learns OCC boundaries for each cluster. On the right column, the incoming chunk is shown, where the previously learned prototypes are utilized to identify the two clusters, and the learned OCC boundaries are applied to detect drift. $\theta_i$ represents the proportion of out-of-distribution samples in the $i$-th cluster of the incoming chunk, $\theta_i > \gamma $ indicates that the specific cluster occurs drift.}
\label{fig_case}
\end{figure}

To further demonstrate the practical applicability and interpretability of our proposed method, a demonstrative example is provided based on a real-world Climate dataset~\cite{kaggle-dataset},
which contains real distributional drift, making it suitable for validating our method’s ability to detect imbalanced concept drift. As shown in Fig.~\ref{fig_case}, we visualize both the base and incoming data chunks. Initially, the base chunk is divided into small clusters using fine-grained prototypes, which are subsequently fused into larger, interpretable clusters (i.e., concepts). One-cluster classifiers (OCCs) are independently trained on these clusters to learn their respective OCC boundaries. When the incoming chunk arrives, ICD3 captures the distribution of each concept based on the previously learned structure. Our method then detects drift at each of the clusters. As shown in the Figure, the observed proportion of out-of-distribution samples for cluster 2 reaches 0.475, exceeding the predefined threshold of 0.2, thus identifying a drift alarm. This example demonstrates the interpretability and effectiveness of our approach in complex dynamic real environments.

\subsection{Parameter Sensitivity Study}
\label{section4: pae}
The parameter $\gamma$ represents the threshold for the proportion of out-of-distribution data in the overall number of samples within a cluster. If $\gamma$ is set too low, ICD3 may be over-sensitive to the slight perturbations, which typically do not change the concept distribution, and thus usually not considered as drift.
Conversely, if $\gamma$ is set too high, ICD3 could become overly conservative when issuing drift warnings, thereby overlooking some gradual or slight drifts. So, we analyze the Accuracy of ICD3 on different datasets under $\gamma=\{0.1,0.2,0.3,0.4,0.5,0.6,0.8 \}$, and the results are presented in Table~\ref{tb_p_stu}. It shows the stability of ICD3 against different values of $\gamma$, and it can be observed that when $\gamma$ is set between 0.2 and 0.4, the performance of MICD3 reaches the optimum. Since MICD3 achieves the highest best-performing frequency across the 14 datasets when $\gamma=0.2$, we suggest such a setting and also adopt it on all the datasets without manual tuning in our experiments.

\begin{table}[!t]
\caption{Parameter sensitivity experiments are conducted to evaluate the impact of threshold $\gamma$ on the Accuracy metric. 
}
\label{tb_p_stu}
\centering
\resizebox{1\columnwidth}{!}{
\begin{tabular}{l|ccccccc}
\toprule
Dataset &0.1 &0.2 &0.3 &0.4 &0.5 &0.6 &0.8 \\ 
\midrule
2D-2G-M &0.634 &\textbf{0.766} &0.688 &0.576 &0.510 &0.500 &0.500 \\
2D-2G-C &0.560 &\textbf{0.932} &0.902 &0.822 &0.664 &0.558 &0.502 \\
2D-2G-V &0.592 &0.910 &\textbf{0.928} &0.848 &0.750 &0.662 &0.508 \\
2D-4G-M &0.504 &0.610 &\textbf{0.676} &0.656 &0.578 &0.520 &0.500 \\
2D-4G-C &0.514 &0.796 &\textbf{0.882} &0.778 &0.654 &0.548 &0.500 \\
2D-8G-M &0.504 &\textbf{0.751} &0.670 &0.594 &0.542 &0.514 &0.502 \\
2D-M-N &0.652 &\textbf{0.918} &0.848 &0.724 &0.622 &0.558 &0.502 \\
Avila &0.514 &\textbf{0.550} &0.540 &0.526 &0.504 &0.500 &0.500 \\
Covtype &0.494 &\textbf{0.517} &0.500 &0.498 &0.498 &0.500 &0.500 \\
Insect &0.512 &\textbf{0.534} &0.522 &0.514 &0.500 &0.500 &0.500 \\
Noaa &0.500 &0.776 &\textbf{0.875} &0.630 &0.516 &0.500 &0.500 \\
Occupy &0.504 &\textbf{0.805} &0.768 &0.582 &0.510 &0.504 &0.500 \\
Posture &0.500 &0.506 &\textbf{0.511} &0.492 &0.496 &0.500 &0.500 \\
Shuttle &0.556 &0.785 &0.856 &\textbf{0.864} &0.718 &0.574 &0.500 \\ \midrule
Ave. Rank &5.179 &\textbf{1.571} &1.714 &3.321 &4.607 &5.321 &6.286 \\ 
\bottomrule
\end{tabular}
}
\end{table}

\section{Conclusion}

This paper proposes the Imbalanced Cluster Descriptor-based Drift Detection (ICD3) method to detect, understand, and interpret drifts that occur on imbalanced concepts in a challenging unsupervised environment.
Differing from most existing discriminative counterparts that concentrate on the detection of drifts from the perspective of a whole data chunk, ICD3 finely describes each concept with robustness to their imbalanced scales, and monitors their individual changing. 
More specifically, ICD3 adopts multiple independent One-Cluster Classifiers (OCCs) to describe and track drift within each cluster, thus circumventing the risk of overlooking small concepts. In addition, the designed drift monitoring mechanism can informatively reveal: 1) ``whether drift occurs'' according to the classification performance of each OCC, 2) ``where the drifts are'' through the corresponding prototypes, and 3) ``how the drifted regions look like'' by interpreting the detected drifted samples with the corresponding concept.
It turns out that ICD3 is competent in both concept drift detection and drift understanding, providing insights for advancing from drift detection to drift understanding and adaptation in unsupervised and imbalanced contexts.
Comprehensive experiments show the promising characteristics of ICD3.

\printbibliography

\begin{IEEEbiography}[{\includegraphics[width=1in,height=1.25in,keepaspectratio]{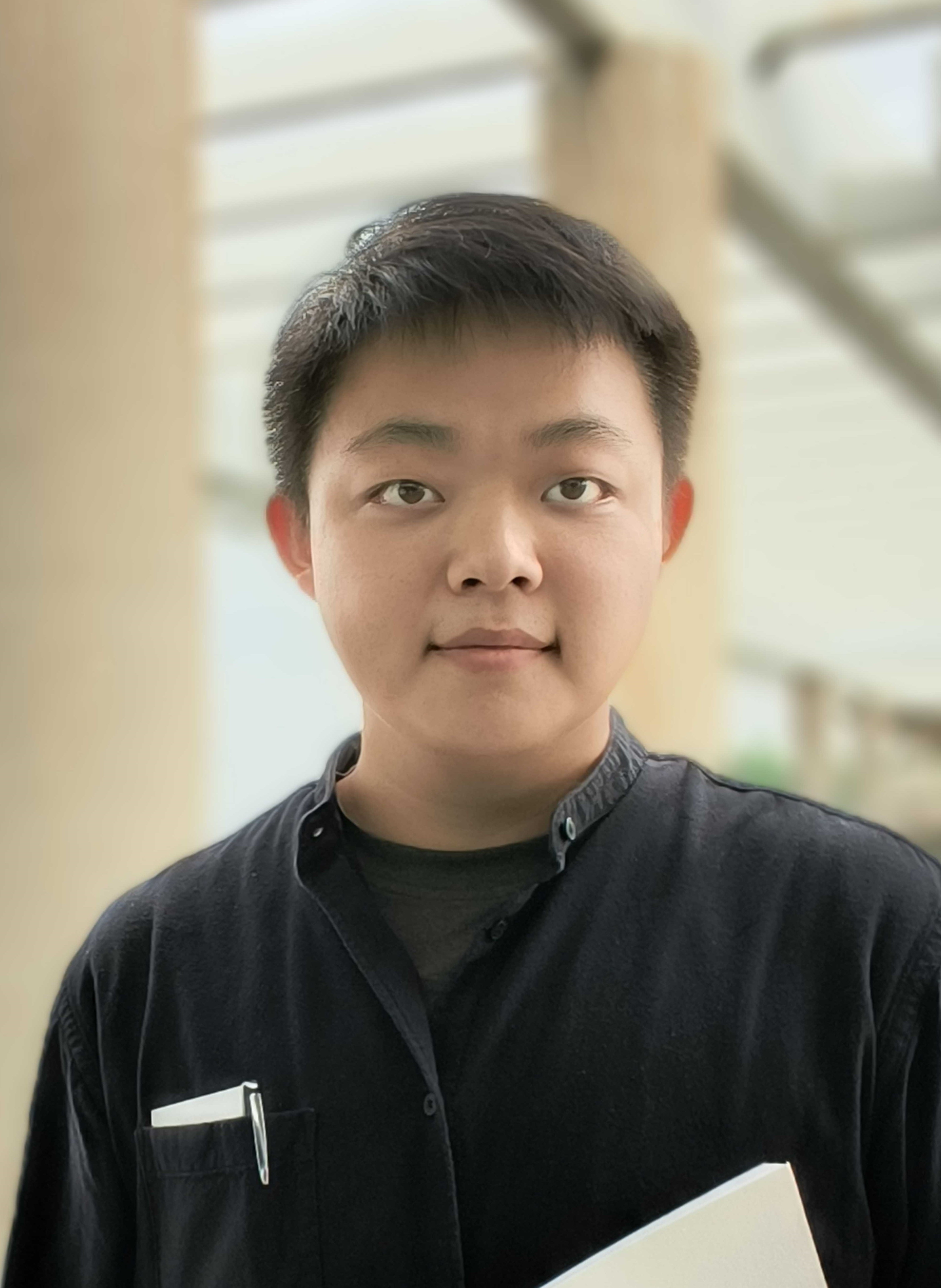}}]{Yiqun Zhang} (Senior Member, IEEE) received the B.Eng. degree from South China University of Technology in 2013, and the M.S. and Ph.D. degrees from Hong Kong Baptist University in 2014 and 2019, respectively. He is currently an Associate Professor at the School of Computer Science and Technology, Guangdong University of Technology, and a Visiting Research Scholar at the Department of Computer Science, Hong Kong Baptist University. 
His current research interests include machine learning, data mining, and their applications. 
\end{IEEEbiography}

\begin{IEEEbiography}[{\includegraphics[width=1in,height=1.25in,keepaspectratio]{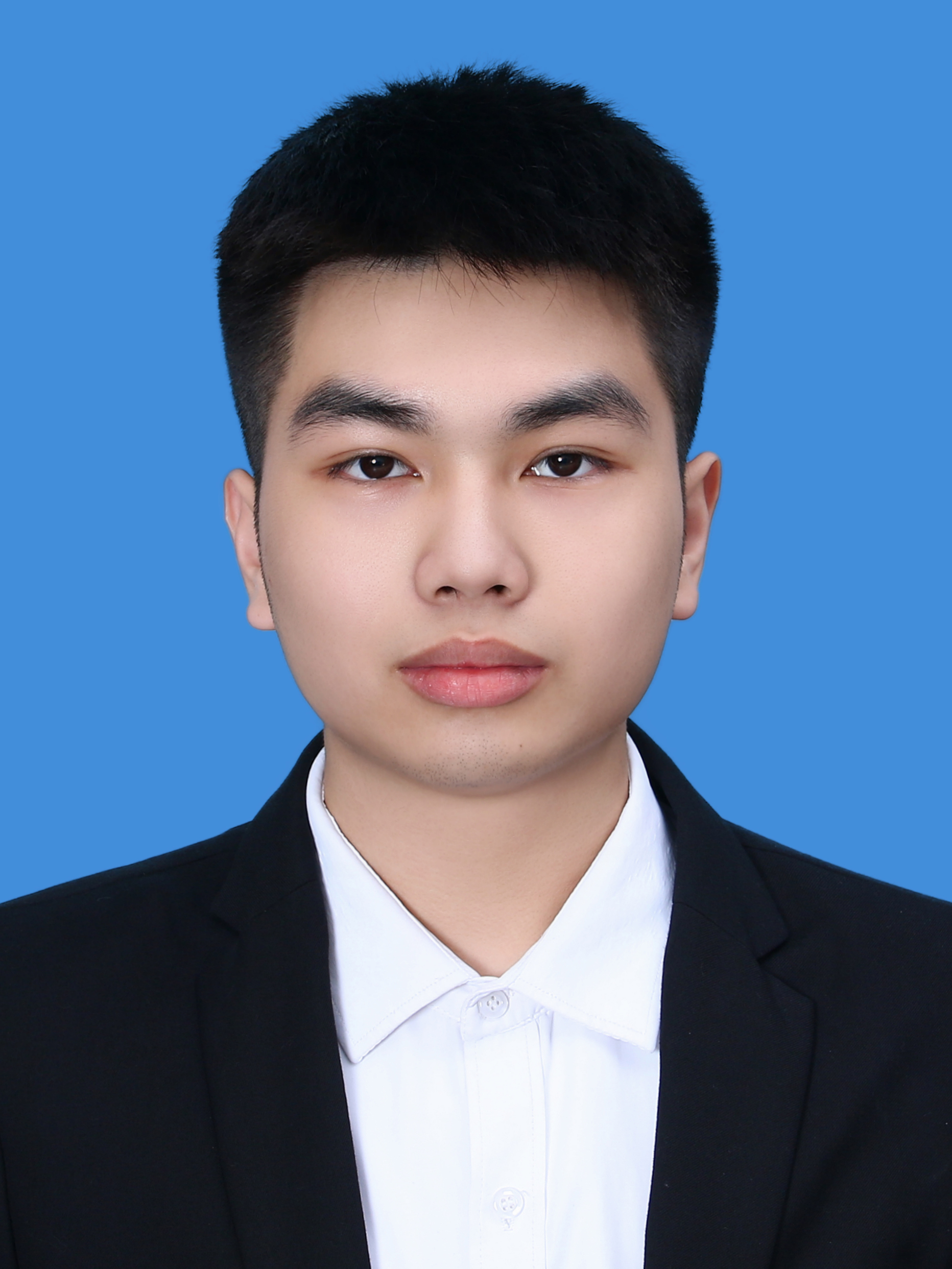}}]{Zhanpei Huang}
 received the B.S. degree from Guangdong University of Technology, Guangdong, China, in 2024. He is currently a graduate student at Guangdong University of Technology, Guangzhou, China. His current research interests include concept drift detection, self-supervised learning, and machine learning.
\end{IEEEbiography}

\begin{IEEEbiography}[{\includegraphics[width=1in,height=1.25in,keepaspectratio]{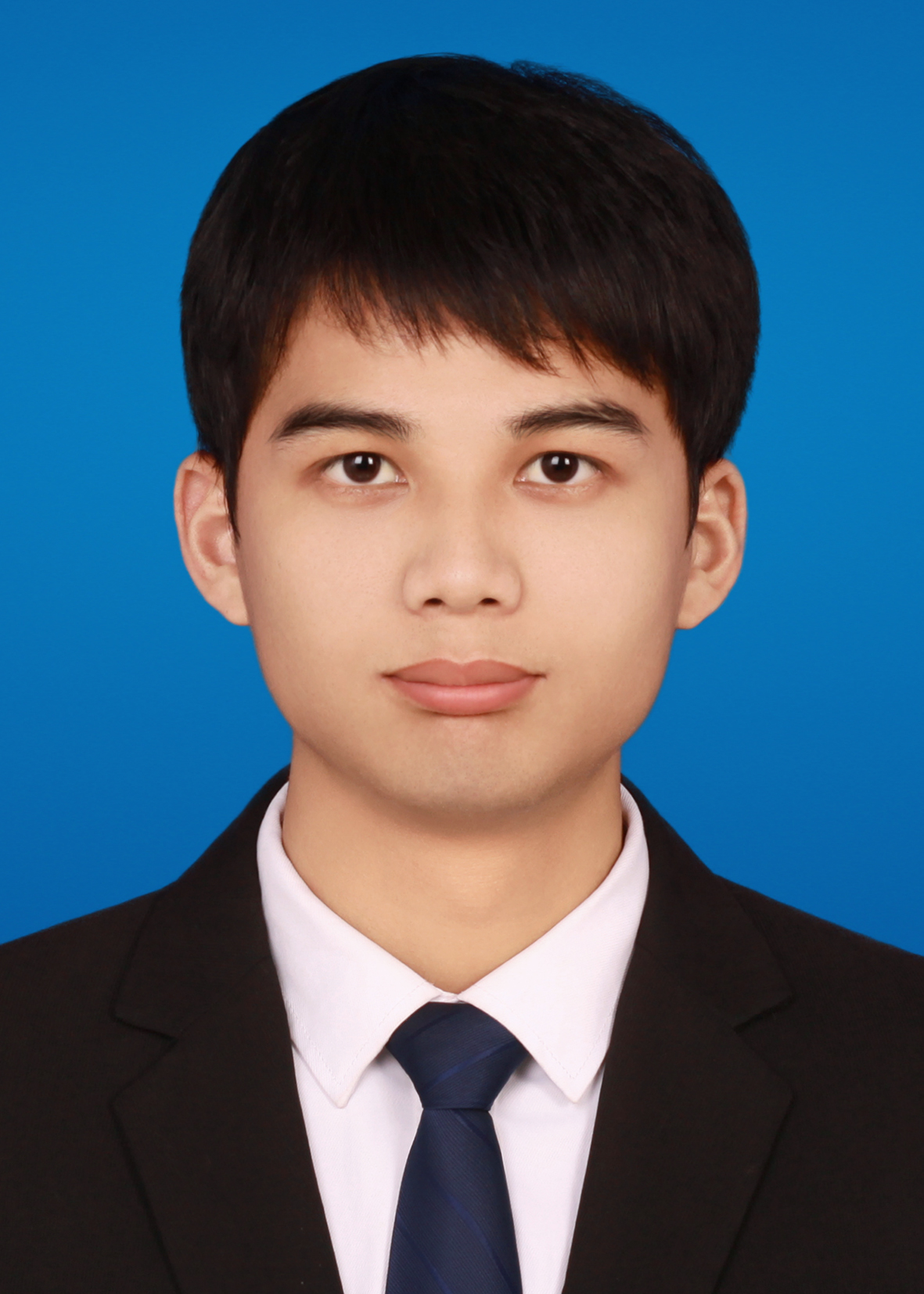}}]{Mingjie Zhao}
 received the B.Eng. degree in software engineering from East China University of Technology, Jiangxi, China, in 2022, and the M.S. degree from Guangdong University of Technology, Guangzhou, China, in 2025. He is now pursuing a Ph.D. degree at Hong Kong Baptist University, Hong Kong, China. His current research interests include concept drift detection, data mining, and machine learning on imbalanced data.
\end{IEEEbiography}

\begin{IEEEbiography}[{\includegraphics[width=1in,height=1.25in,keepaspectratio]{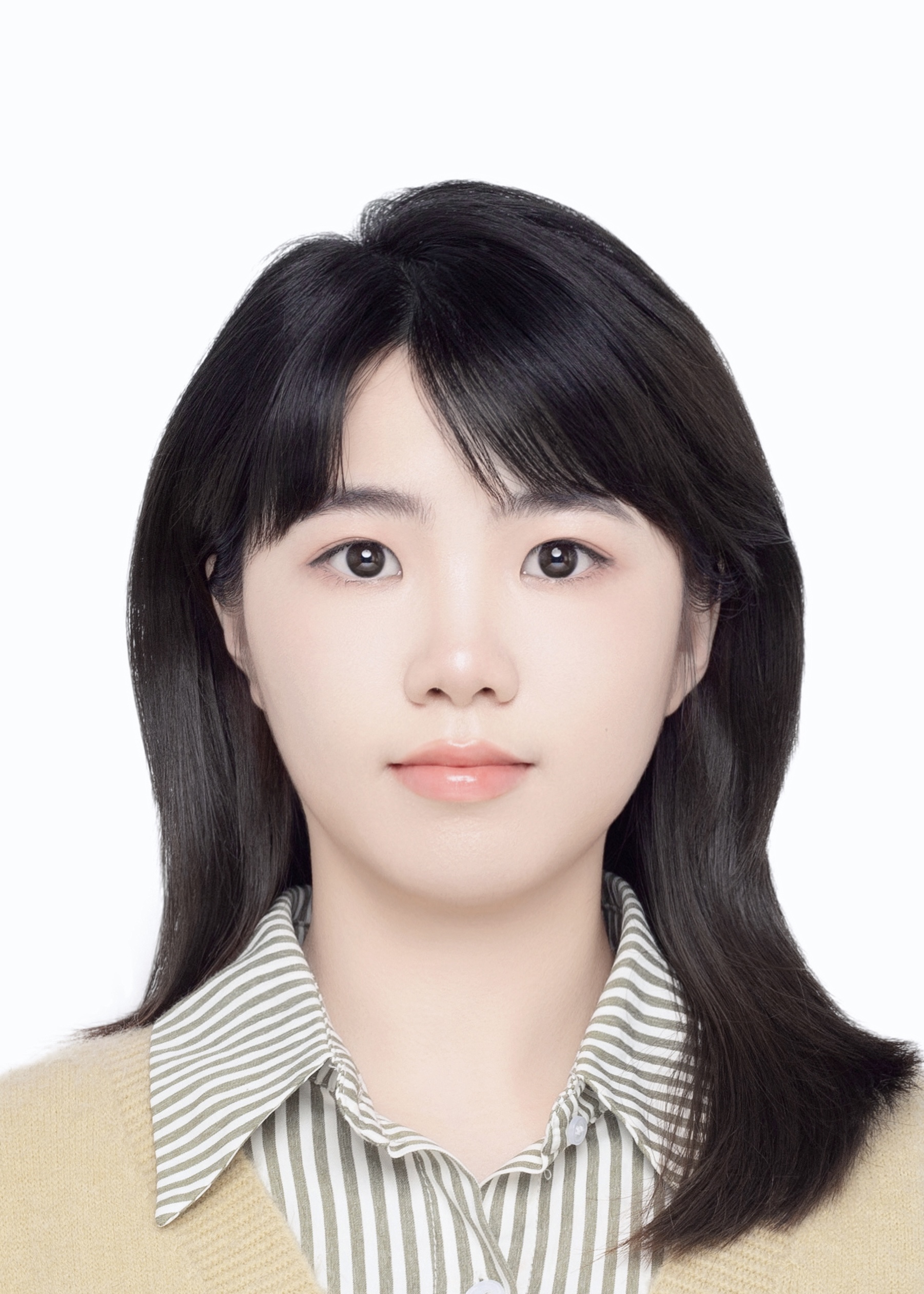}}]{Chuyao Zhang}
 received the B.Eng. degree in software engineering from Guangdong University of Finance, Guangzhou, China, in 2023. She is currently a graduate student at Guangdong University of Technology, Guangzhou, China. Her current research interests include missing value completion, data mining, and machine learning.
\end{IEEEbiography}

\begin{IEEEbiography}[{\includegraphics[width=1in,height=1.25in,keepaspectratio]{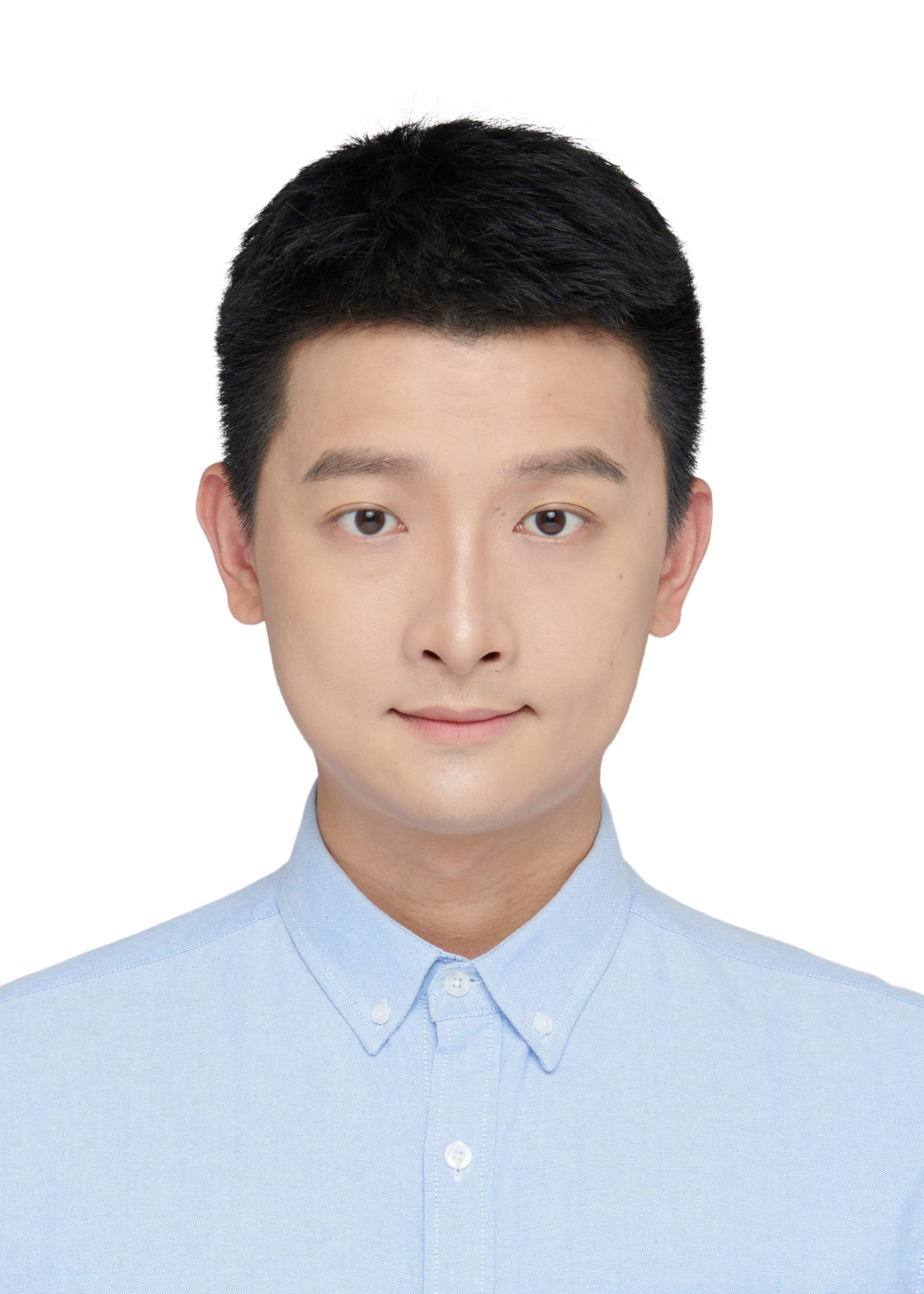}}]{Yang Lu} (Senior Member, IEEE)
received the B.S. and M.S. degrees in software engineering from the University of Macau, Macau, China, in 2012 and 2014, respectively, and the Ph.D. degree in computer science from Hong Kong Baptist University, Hong Kong, China, in 2019. He is currently an Assistant Professor with the Department of Computer Science and Technology, School of Informatics, Xiamen University, Xiamen, China. His current research interests include open-world robust deep learning, such as long-tail learning, federated learning, label-noise learning, and continual learning.
 \end{IEEEbiography}

\begin{IEEEbiography}[{\includegraphics[width=1in,height=1.25in,keepaspectratio]{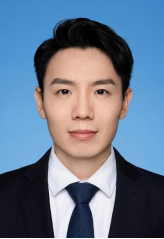}}]{Yuzhu Ji} (Member, IEEE) received a B.S. degree in computer science from the PLA Information Engineering University, Zhengzhou, China, in 2012, and the M.S. and Ph.D. degrees from the Department of Computer Science, Harbin Institute of Technology Shenzhen, China, in 2015 and 2019. He is currently an Associate Professor at the School of Computer Science and Technology, Guangdong University of Technology, Guangzhou, China. His current research interests include salient object detection, image segmentation, and graph clustering.
\end{IEEEbiography}

\begin{IEEEbiography}[{\includegraphics[width=1in,height=1.25in,keepaspectratio]{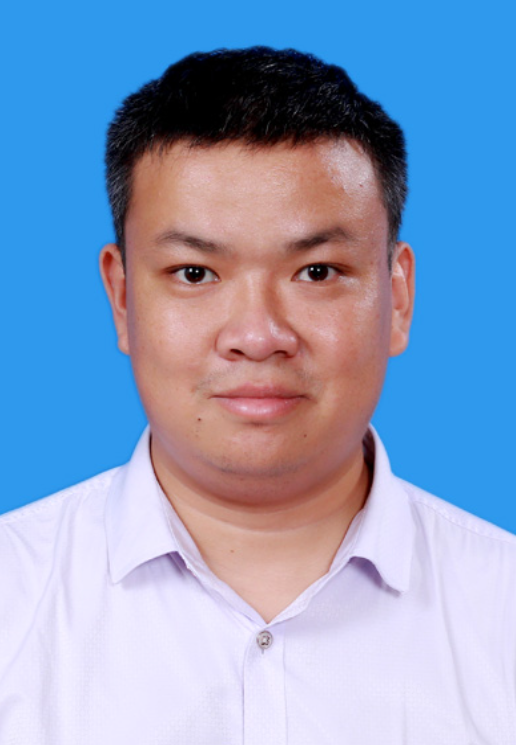}}]{Fangqing Gu} received the B.S. degree from Changchun University, Jilin, China, in 2007, the M.S. degree from the Guangdong University of Technology, Guangzhou, Guangdong, China, in 2011, and the Ph.D. degree from the Department of Computer Science, Hong Kong Baptist University, Hong Kong, in 2016.
He is currently an associate professor at the School of Mathematics and Statistics of Guangdong University of Technology. His current research interests include data mining, machine learning, and evolutionary computation.
\end{IEEEbiography}

\begin{IEEEbiography}[{\includegraphics[width=1in,height=1.25in,keepaspectratio]{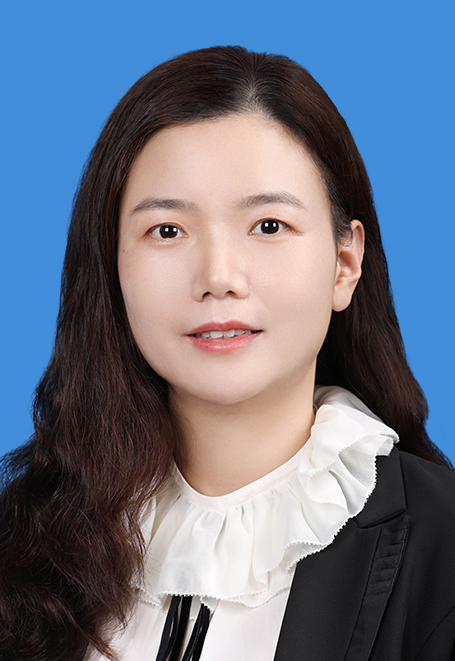}}]{An Zeng} (Member, IEEE)  received the B.S. degree from Nanchang Hangkong University, Nanchang, China, in 1999, and the M.S. and Ph.D. degrees in applied computer technology from South China University of Technology, Guangzhou, China, in 2002 and 2005, respectively. From 2008 to 2010, she was a Postdoctoral Fellow with the Faculty of Computer Science, Medical School of Dalhousie University, Halifax, NS, Canada. She is currently a Professor and the Vice Dean of the School of Computer Science and Technology, Guangdong University of Technology, Guangzhou. Her research interests include deep learning, artificial intelligence, and their applications in the fields of medicine and healthcare.
\end{IEEEbiography}

\end{document}